\documentclass[manuscript]{acmart}
\AtBeginDocument{%
  }

\setcopyright{acmlicensed}
\copyrightyear{2018}
\acmYear{2018}
\acmDOI{XXXXXXX.XXXXXXX}
\acmConference[Conference acronym 'XX]{Make sure to enter the correct
  conference title from your rights confirmation email}{June 03--05,
  2018}{Woodstock, NY}
\acmISBN{978-1-4503-XXXX-X/2018/06}

\usepackage{subcaption}
\usepackage{hyperref}
\usepackage{multirow}
\usepackage{array}
\begin{document}

\title{Towards Knowledge Guided Pretraining Approaches for Multimodal Foundation Models: Applications in Remote Sensing}

\author{Praveen Ravirathinam}
\email{pravirat@umn.edu}
\affiliation{%
  \institution{University of Minnesota}
  \city{Twin Cities}
  \state{Minnesota}
  \country{USA}
}

\author{Ajitesh Parthasarathy}
\email{parth057@umn.edu}
\affiliation{%
  \institution{University of Minnesota}
  \city{Minneapolis}
  \state{MN}
  \country{USA}
}

\author{Ankush Khandelwal}
\email{khand035@umn.edu}
\affiliation{%
  \institution{University of Minnesota}
  \city{Minneapolis}
  \state{MN}
  \country{USA}
}

\author{Rahul Ghosh}
\email{ghosh128@umn.edu}
\affiliation{%
  \institution{University of Minnesota}
  \city{Minneapolis}
  \state{MN}
  \country{USA}
}

\author{Vipin Kumar}
\email{kumar001@umn.edu}
\affiliation{%
  \institution{University of Minnesota}
  \city{Minneapolis}
  \state{MN}
  \country{USA}
}

\renewcommand{\shortauthors}{Ravirathinam et al.}

\begin{abstract}
    Self-supervised learning has emerged as a powerful paradigm for pretraining foundation models using large-scale data. Existing pretraining approaches predominantly rely on masked reconstruction or next-token prediction strategies, demonstrating strong performance across various downstream tasks, including geoscience applications. However, these approaches do not fully capture the knowledge of causal interplay between different geospatial and environmental variables. To address this limitation, we propose Knowledge Guided Variable-Step Forecasting (KG-VSF), a novel pretraining task that models forecasting as a conditional generation task, where driver variables (e.g., weather) inform the prediction of response variables (e.g., satellite imagery). We demonstrate that pretraining in such a fashion leads to strong embeddings which give enhanced performance when finetuned on downstream tasks where capturing this causality matters such as pixel wise crop type mapping, soil moisture estimation and forecasting, missing image prediction, and future image forecasting when compared to finetuning embeddings from other standard pretraining approaches. 
    
\end{abstract}

\ccsdesc[300]{Computing methodologies~Computer vision}
\ccsdesc[500]{Applied computing~Earth and atmospheric sciences}
\ccsdesc[500]{Computing methodologies~Unsupervised learning}
\ccsdesc[300]{Computing methodologies~Neural networks}
\ccsdesc[300]{Computing methodologies~Knowledge representation and reasoning}

\keywords{Pretraining Tasks, Knowledge Guided, Foundation Model, Remote Sensing}

\received{20 February 2007}
\received[revised]{12 March 2009}
\received[accepted]{5 June 2009}

\maketitle

\section{Introduction}

\begin{figure*}[t]
    \centering
    \begin{minipage}{0.98\textwidth} 
        \centering
        \begin{subfigure}{0.33\textwidth} 
            \centering
            \begin{subfigure}{\textwidth}
                \centering
                \includegraphics[height=1cm]{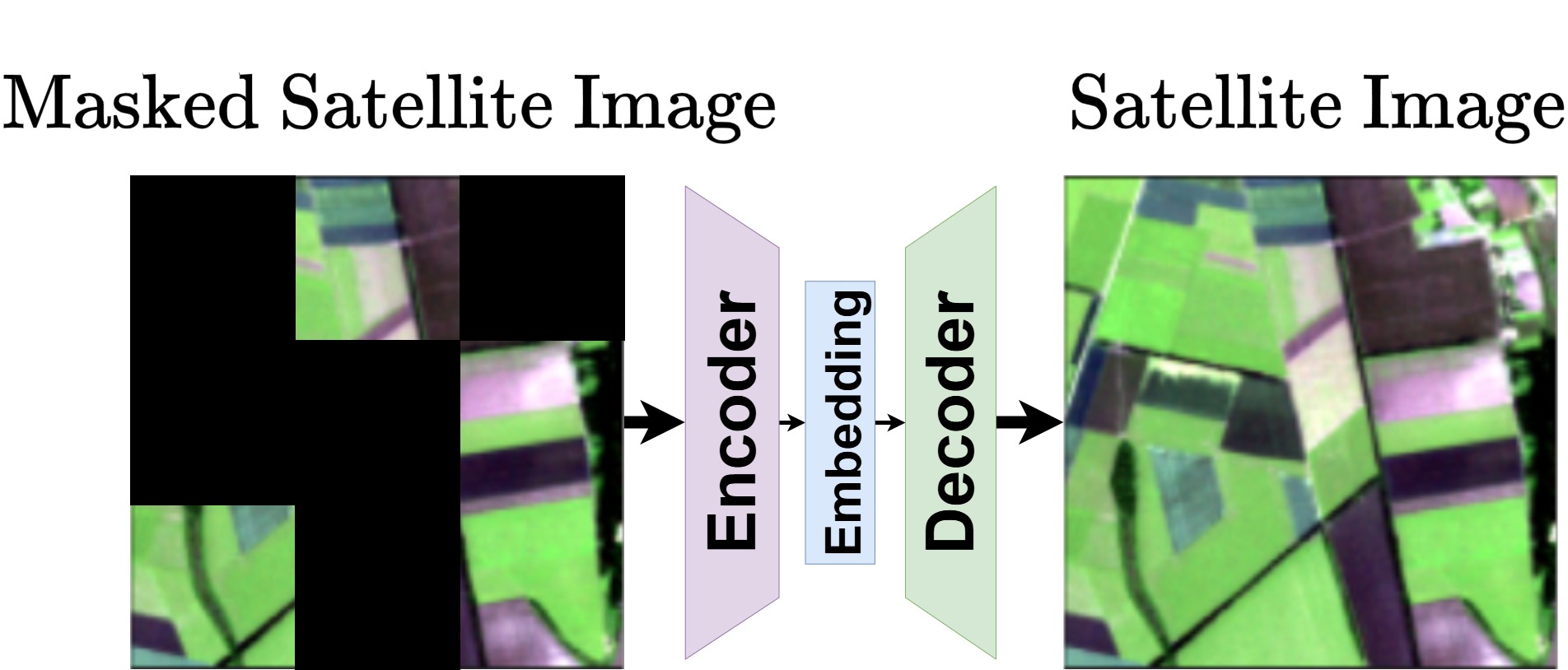}
                \caption{Single-Modality Masked Reconstruction (SM-MR)}
                \label{fig:smmae_pretrain_task}
            \end{subfigure}
            \vfill
            \begin{subfigure}{\textwidth}
                \centering
                \includegraphics[height=2cm]{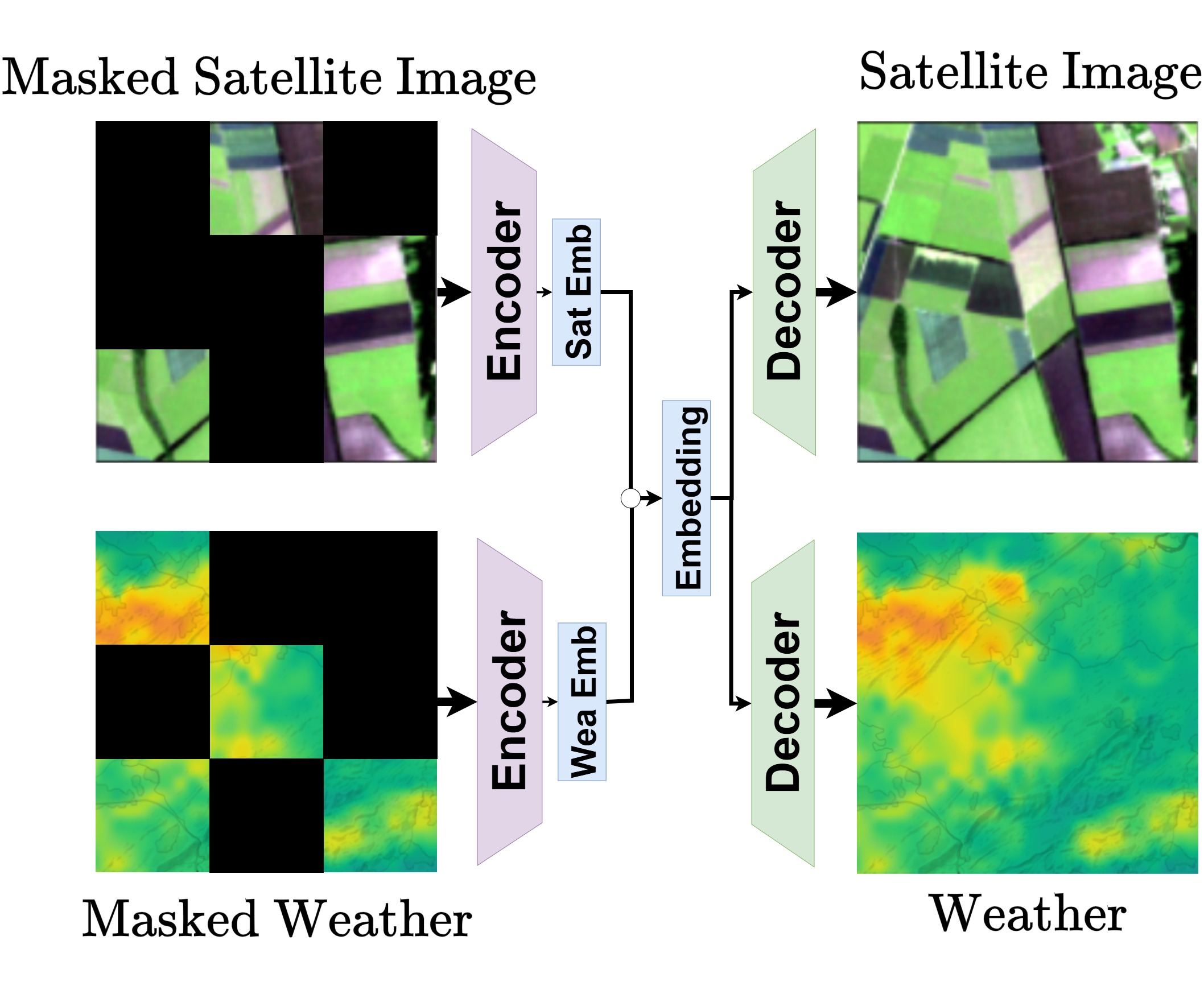}
                \caption{Multi-Modality Masked Reconstruction (MM-MR)}
                \label{fig:mmmae_pretrain_task}
            \end{subfigure}
        \end{subfigure}
        \hfill
        \begin{subfigure}{0.31\textwidth}
            \centering
            \includegraphics[height=4.3cm]{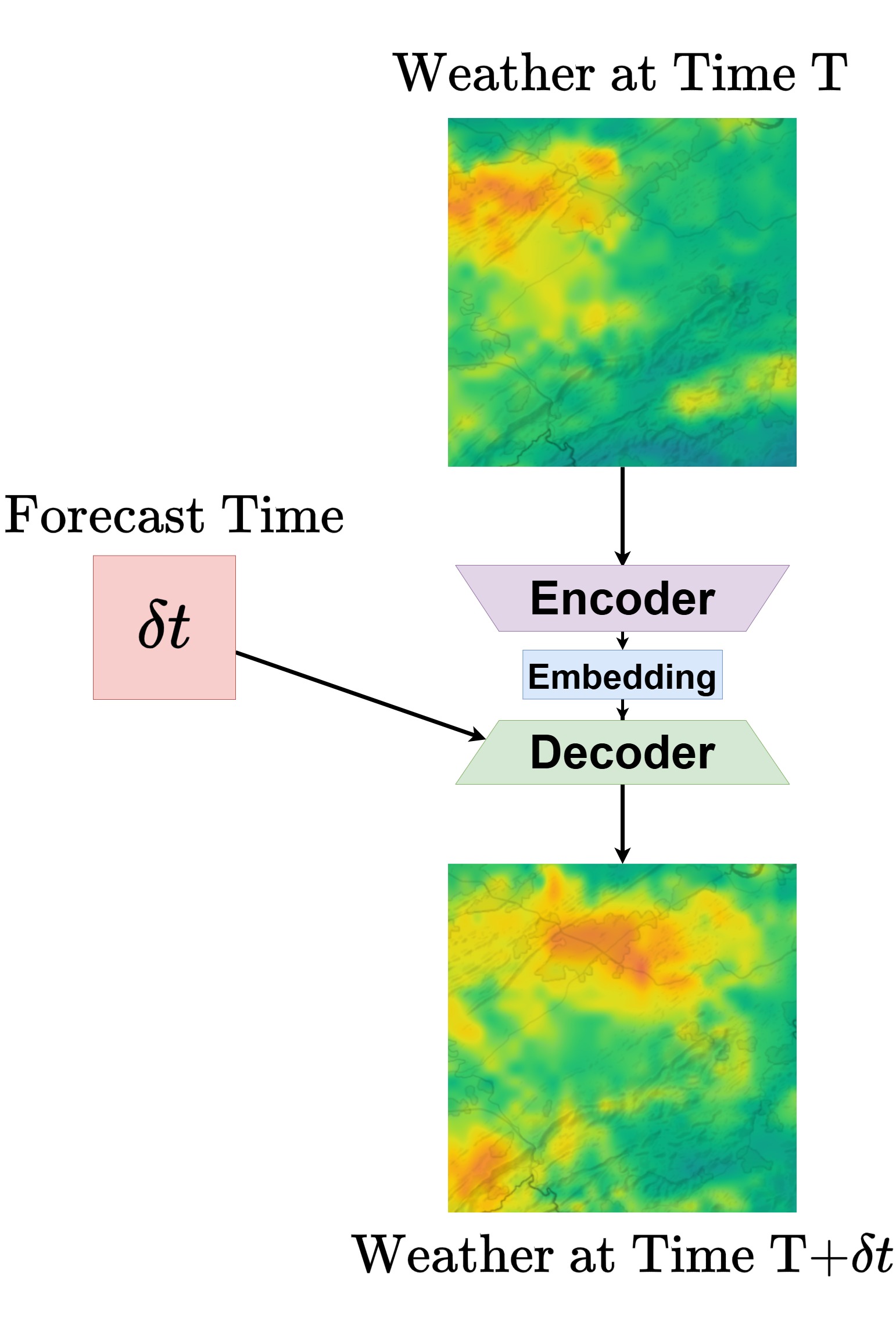}
            \caption{Single Modal Variable Step Forecasting (SM-VSF)}
            \label{fig:smvsf_pretrain_task}
        \end{subfigure}
        \hfill
        \begin{subfigure}{0.31\textwidth}
            \centering
            \includegraphics[height=4.3cm]{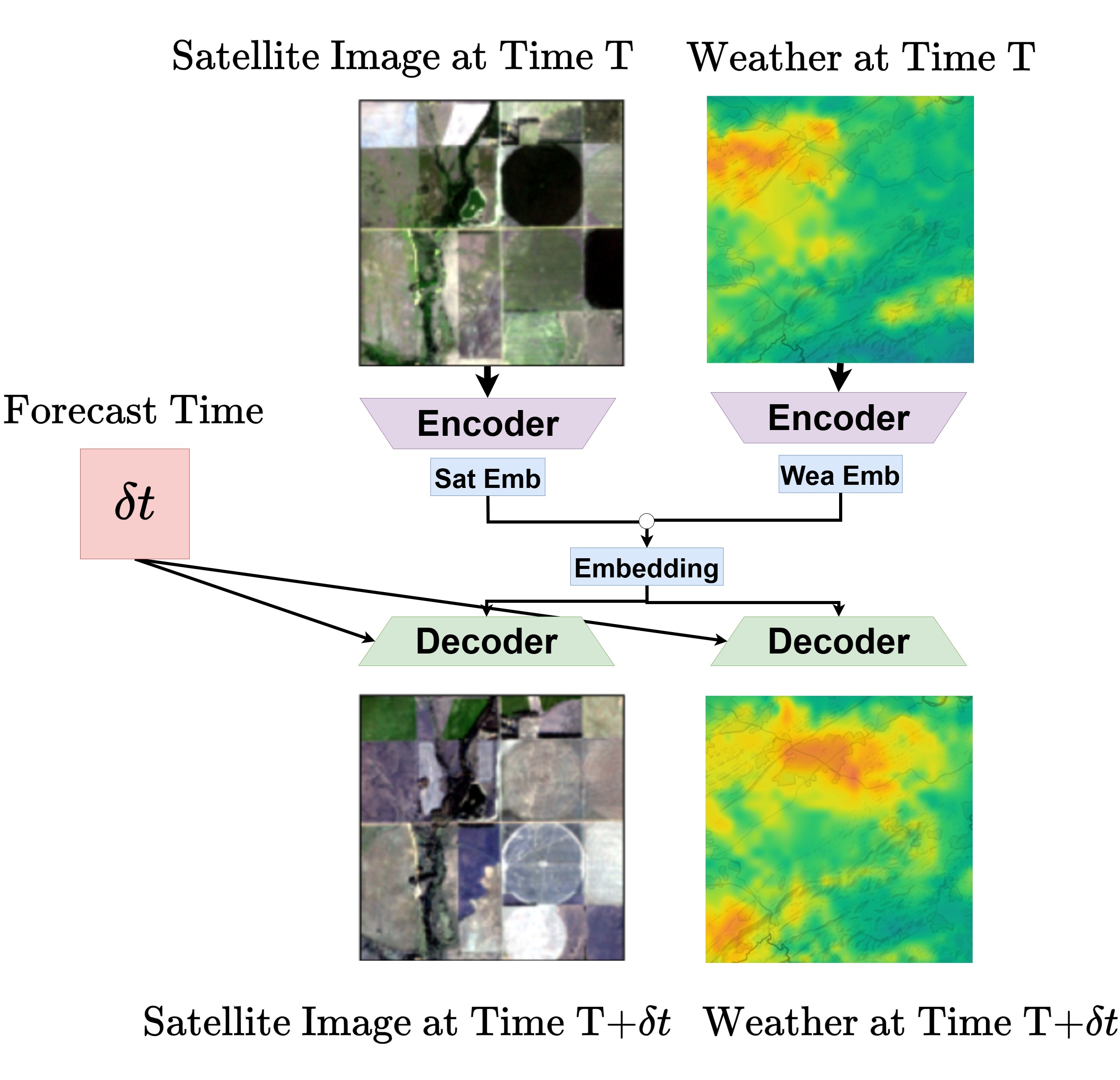}
            \caption{Multi Modal Variable Step Forecasting (MM-VSF)}
            \label{fig:mmvsf_pretrain_task_old}
        \end{subfigure}
        \vspace{-0.35cm}
        \caption{\small{Comparison of different pretraining tasks: Masked Reconstruction (MR) and Variable Step Forecasting (VSF) in both single and multi-modality settings. Please zoom for better viewing.}}
        \label{fig:pretraining_tasks}
        \vspace{-0.35cm}
    \end{minipage}
\end{figure*}

\begin{figure*}[t]
    \centering
    \includegraphics[height=4cm]{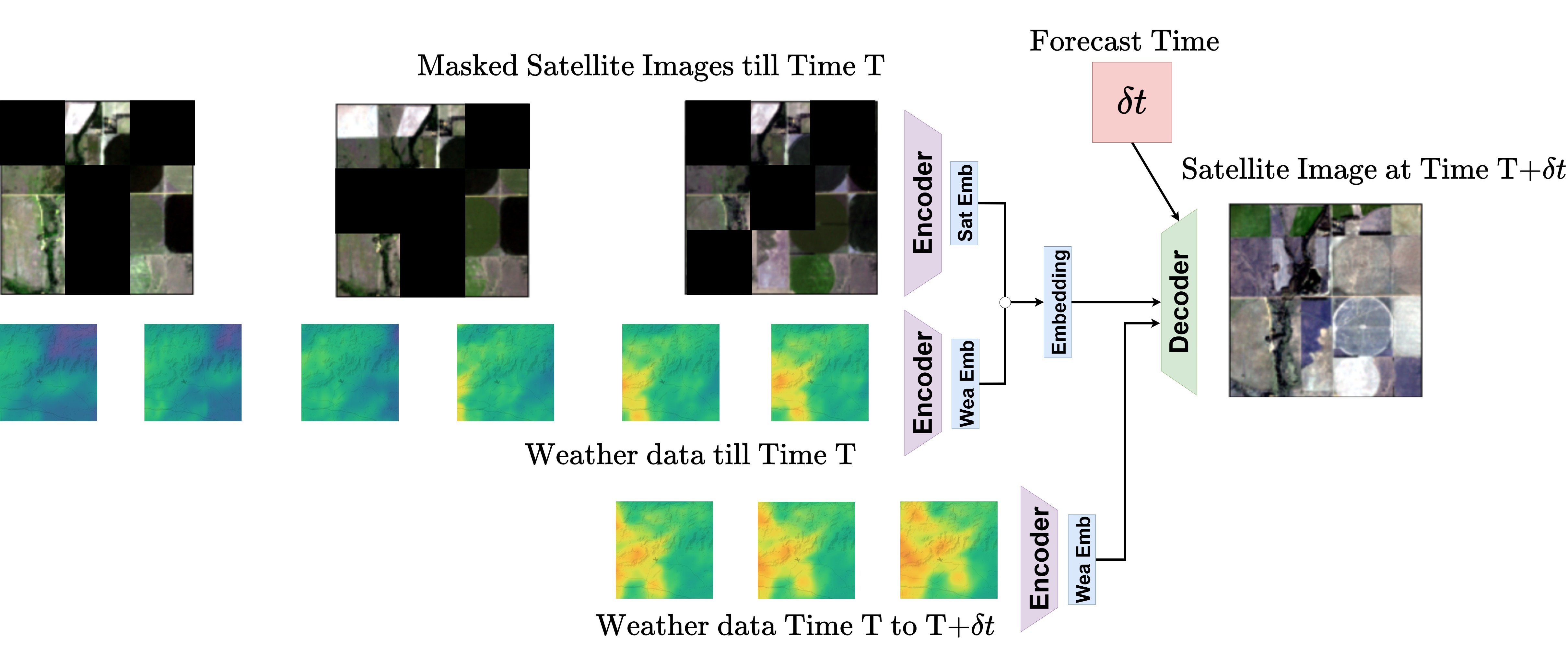}
    \vspace{-0.35cm}
    \caption{\small{Our proposed novel Knowledge Guided Variable Step Forecasting (KG-VSF) task. Our pretraining task estimates a satellite image in the future (satellite data at Time $T+\delta t$) using satellite imagery and weather context (satellite and weather data till Time $T$) and weather data up to that future date (weather data from Time $T\ to\ T+\delta t$).}}
    
    \label{fig:mmvsf_pretrain_task}
\end{figure*}

Increased availability and ease of access to large scale satellite data has enabled the development of deep learning models for diverse geoscience applications, including land cover mapping~\cite{statt,kussul2017deep}, wildfire mapping~\cite{nayak2018classifying,seydi2022burnt,zhao2018saliency}, crop yield prediction~\cite{kuwata2015estimating,you2017deep}, flood forecasting~\cite{bentivoglio2022deep} etc. This has also led to the emergence of a large number of geoscience foundation models\cite{jakubik2023foundationmodelsgeneralistgeospatial,cha2023billion,mall2023remote,deng2024k2,hong2024spectralgpt,guo2024skysense,mai2022towards,bastani2023satlaspretrain,lu2024ai,schmude2024prithviwxcfoundationmodel,liu2024remoteclip}.

Foundation models are a new class of machine learning models that are pre-trained using self-supervised learning approaches on large-scale datasets to generate high-quality embeddings\cite{gui2024survey,jing2020self}. These embeddings, after fine-tuning, have demonstrated strong  performance across various downstream tasks \cite{awais2025foundation,liang2024foundation,zhou2024comprehensive,li2024foundation}. There are two broad categories of pretraining tasks that are used in most of the existing foundation models. The first category includes variants of masked reconstruction ~\cite{he2022masked,devlin2018bert} and next-token prediction\cite{achiam2023gpt} that focus on reconstruction or sequence forecasting.  The second category includes contrastive learning and distillation \cite{chen2020simpleframeworkcontrastivelearningsimclr,caron2021emergingpropertiesselfsupervisedvisiondino}, which focus on aligning embeddings across modalities.  Our work in this paper builds on the first category by introducing a new pretraining task for use geospatial foundation models that leverages causal relationships among modalities; i.e. some variables (drivers) directly influence other variables (responses). The distinction from contrastive/distillation based approaches lies in the type of relationships being modeled, which will be revisited later for context.

In masked reconstruction (MR)-based pretraining, portions of the input data are hidden (masked), and the model is trained to encode these masked input data into an embedding that is used to reconstruct the original (unmasked) data. This style of pretraining has shown great progress in the vision domain (where parts of the image are masked\cite{he2022masked}) and in the language domain (where words from a sentence are masked\cite{devlin2018bert}). In the geoscience domain, masked reconstruction has been used extensively in the building of remote sensing foundation models \cite{jakubik2023foundationmodelsgeneralistgeospatial,cha2023billion,cong2022satmae,tseng2023lightweight,lu2024ai}. In such models, parts of the satellite image are masked, and the pretraining objective is to reconstruct the entire satellite image (denoted by Figure \ref{fig:smmae_pretrain_task}). Utility of embeddings produced via a single-modality masked reconstruction (SM-MR) pre-training objective has been demonstrated for numerous tasks where the aim is to estimate or infer current or past variables using available data. Examples of such tasks include flood inundation mapping, wildfire scar mapping, cloud removal, urban semantic segmentation mapping, scene classification etc. \cite{jakubik2023foundationmodelsgeneralistgeospatial, cong2022satmae, sun2022ringmo}. Variations of this pretraining task have been developed, where the pre-training task involves masking multiple modalities (MM) and reconstructing them (see Figure \ref{fig:mmmae_pretrain_task}), where instead of one modality, multiple masked modalities would be used. Embeddings produced by such a pre-training task (MM-MR - see Figure \ref{fig:mmmae_pretrain_task}) have been shown to lead to improvements in scene classification and crop field classification \cite{nedungadi2024mmearth,tseng2023lightweight}. 

In next-token prediction-based pretraining, the model learns to predict the next or subsequent data points in a given sequence. This style of pretraining has been used in language modeling (where given a sequence of words the model learns to predict the next word\cite{touvron2023llama}) most notably GPT\cite{achiam2023gpt}. In the geoscience domain, next token prediction style of pretraining has been used in the climate/weather foundation models \cite{nguyen2023climax,schmude2024prithviwxcfoundationmodel,pathak2022fourcastnet,gao2022earthformer}. Here, given a series of masked or unmasked climate variables, the objective is to estimate these climate variables at a random future timestamp (see Figure \ref{fig:smvsf_pretrain_task}). This pretraining task has been augmented to include multiple modalities where instead of trying to forecast one variable, future values of multiple variables are estimated from a common embedding (see Figure \ref{fig:mmvsf_pretrain_task_old}) Models pretrained in this variable step forecasting (VSF) fashion have shown great performance in tasks where the aim is to predict future variables based on past observations. Examples of such tasks include precipitation forecasting, hurricane trajectory tracking, and sea surface temperature prediction.

In geoscience applications, there is often a complex interplay between different variables i.e., one set of variables (drivers) influence the other set of variables (response). For example, weather (driver) influences the vegetation growth (response) which can be observed by a satellite. Note that the response observed by satellite in response to weather drivers will differ depending on the state/nature of the location.  For example, grasslands may green up after rain, evergreen forests and urban areas may not change much in response to the rain, and different types of crops may be impacted by weather differently.  Neither MR nor VSF  attempt to capture  such causal relationships between these sets of variables for learning robust embeddings.  MR and VSF based pretraining primarily capture structure in the data in space and time, respectively, and do not explicitly leverage knowledge of causal relationship across modalities. 

To address this gap, we propose Knowledge Guided Variable-Step Forecasting (KG-VSF), a novel pretraining objective that explicitly leverages known causal dependencies between modalities. In this framework, the pre-training task models forecasting as a conditional generation task where conditioning is governed by the driver variables (e.g. weather) and the task is to predict the response variables (e.g. satellite images). Specifically, given a series of weather data and satellite imagery (at irregular intervals) in the past and weather data up till a future date, our pretraining task is to estimate the satellite images in the future (see Figure \ref{fig:mmvsf_pretrain_task}). 
The embeddings produced in this pre-training task implicitly learn about the nature of the location (e.g., grass land vs forests vs specific crops) based only upon past weather and (irregularly available) satellite imagery, as this information determines differential response of weather on the future satellite imagery at that location. Note that past weather and satellite imagery are freely available on a global scale from government  organizations (e.g., NASA, ESA), making this pre-training task useful in many geoscience domains.  

Here, knowledge-guided refers to using domain knowledge of causal dependencies to define the pretraining objective itself. KG-VSF incorporates the understanding that weather variables act as drivers that influence surface conditions observed by satellites, and this directional relationship is explicitly represented in the learning task. This contrasts with prior multimodal fusion approaches, such as Tseng et al. (2023)\cite{tseng2023lightweight}, which combine multiple modalities without encoding directional influence.

Recent multimodal foundation models such as RemoteCLIP\cite{liu2024remoteclip}, SimCLR\cite{chen2020simpleframeworkcontrastivelearningsimclr}, DINO\cite{caron2021emergingpropertiesselfsupervisedvisiondino}, and SkySense\cite{guo2024skysense} have demonstrated impressive success through contrastive or distillation-based objectives that align co-occurring representations across modalities. We discuss these approaches in detail in Section 2, as they represent an important and complementary direction. However, their objectives primarily capture semantic correlations between modalities rather than causal dependencies. In contrast, KG-VSF aims to learn causal dynamics, i.e., how one modality (e.g., weather) drives another (e.g., land surface state).

We evaluate the effectiveness of KG-VSF in settings where the impact of weather on land cover and its properties is visible via satellite imagery.  For this we created 
two data sets (one global and the other for US crop lands), each incorporating Sentinel2 spectral data and ERA5 weather data. These data sets are used  to pre-train models using our proposed KG-VSF method (Figure \ref{fig:mmvsf_pretrain_task}) and the 4 existing pretraining methods (Figure \ref{fig:pretraining_tasks}). We demonstrate that the embeddings produced from our proposed pretraining approach outperform embeddings produced from existing pretraining methods for the following downstream applications: soil moisture related tasks (prediction and forecasting), pixel wise crop type mapping (prediction), and spectral imagery based tasks (prediction and forecasting). We release the code and model used to public (\href{https://drive.google.com/drive/folders/1GwVnjjtuqyso-h_Edoafe-VOmDW2Mpxn?usp=sharing}{Link}). For the extended version of this paper please refer to the Arxiv version (\href{https://arxiv.org/abs/2407.19660}{Link})

The paper is organized as follows: Section 2 reviews related work and highlights how our approach differs from existing research. Section 3 describes the proposed deep learning architecture, and Section 4 details the pretraining task. Section 5 presents the data sources, masking scheme, and sampling strategy for the two datasets. Section 6 discusses the comparative pretraining frameworks, models, and implementation details. Section 7 focuses on the downstream task of crop type mapping, while Section 8 covers the downstream tasks of soil moisture prediction and forecasting. Section 9 examines the downstream tasks of missing-image prediction and future-image forecasting. Section 10 analyzes the quality of embeddings from different pretraining methods. Section 11 explores experiments to verify that KG-VSF captures causal relationships rather than mere correlations. Section 12 evaluates the scalability of the KG-VSF framework. Finally, Section 13 provides discussion and concluding remarks.

\section{Related Work}

Existing geoscience foundation models can be broadly grouped into two categories based on the data used: (1) weather-climate \cite{nguyen2023climax,pathak2022fourcastnet,Bodnar2025Aurora,Bi2023PanguWeather} that are typically used for weather forecasting or climate modeling and (2) spectral data from remote sensing satellites \cite{cong2022satmae,jakubik2023foundationmodelsgeneralistgeospatial,tseng2023lightweight,nedungadi2024mmearth,liu2024remoteclip} that are largely used for identifying land-use land-cover change dynamics. 

The most common pretraining objective in remote sensing models is masked reconstruction of spectral imagery. To improve embedding richness, varying portions of input imagery are masked to make reconstruction harder, leading to stronger representations  \cite{cong2022satmae}. However, embeddings trained this way capture primarily spatial features of individual images and may underperform on tasks that require temporal context, such as crop monitoring or land-cover change detection. To solve this, previous works included multiple timestamps in their input, however some of these methods stacked these images together \cite{jakubik2023foundationmodelsgeneralistgeospatial}, thus removing the temporal aspect. However, some methods added a timestamp positional embedding so that the model has a sense of time \cite{khanna2023diffusionsat,cong2022satmae}. This led to moderate success in handling downstream tasks that require multi temporal contexts.

Another pretraining strategy, forecasting-based pretraining, has been widely used in weather foundation models but less explored for spectral imagery. Models pretrained for masked forecasting have shown success in weather prediction tasks \cite{schmude2024prithviwxcfoundationmodel,pathak2022fourcastnet}. To enhance temporal generalization, variable-future-time forecasting has been introduced, where the forecast horizon varies and is encoded via a delta-time embedding \cite{nguyen2023climax}. Other variants include diffusion-based models, which integrate auxiliary contextual information such as geographic location, time of year, or region \cite{khanna2023diffusionsat}.

A distinct line of work involves contrastive learning and distillation-based approaches such as  RemoteCLIP\cite{liu2024remoteclip}, SimCLR\cite{chen2020simpleframeworkcontrastivelearningsimclr}, DINO\cite{caron2021emergingpropertiesselfsupervisedvisiondino}, and SkySense\cite{guo2024skysense}. Contrastive methods learn by comparing pairs of modalities and maximise similarity between their embeddings and also maximise dissimilarity between the embeddings of unrelated pairs. Distillation-based methods (e.g., DINO\cite{caron2021emergingpropertiesselfsupervisedvisiondino}) instead train a student network to create an embedding that matches a slowly updated embedding produced by the teacher network by using different (augmented) views of the same input.  While these approaches have been highly successful for cross-modal retrieval and semantic understanding, their pretraining objective focuses on co-occurrence alignment rather than modeling causal dependencies between different modalities. In contrast, as noted in the introduction, KG-VSF attempts to explicitly capture the causal impact of one modality over the other (e.g.,  “weather influences the state of the land cover”. Thus, the objective of contrastive/distillation models are complementary but orthogonal to the  objective of KG-VSF, which is designed to capture causal knowledge flow within environmental processes, not semantic similarity across heterogeneous data sources.

\section{Architecture}
\begin{figure*}[t]
  \centering
  \includegraphics[width=0.95\linewidth]{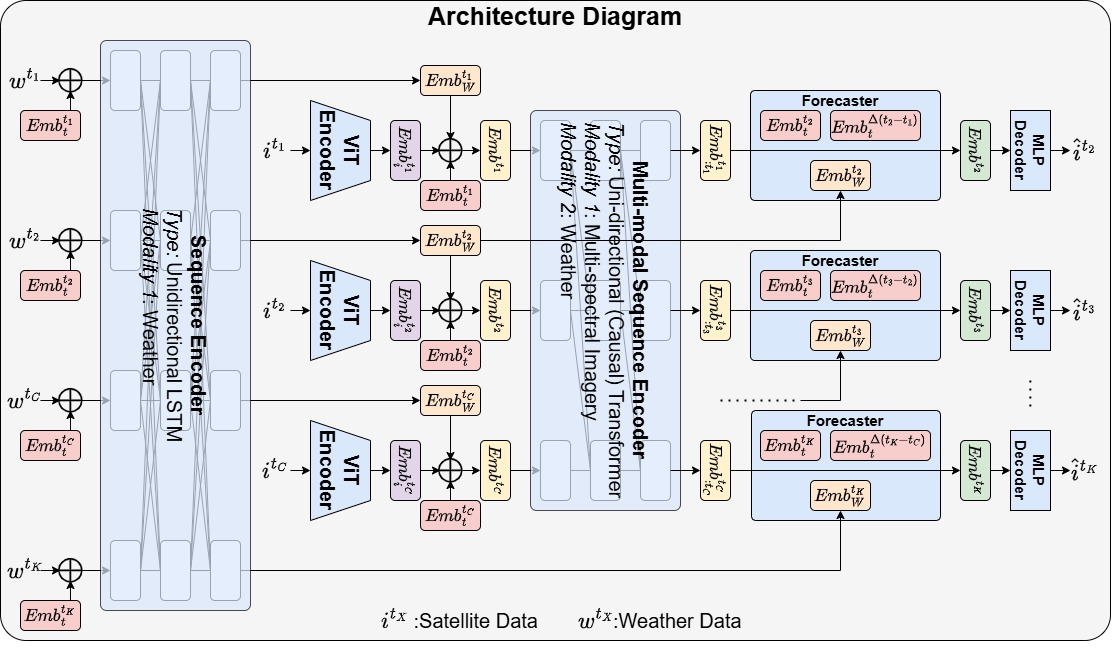}
  \caption{\textit{Knowledge Guided Variable step Forecasting (KG-VSF) Architecture diagram}}
  \label{fig:architecture_detailed}
\end{figure*}

Our architecture follows a heavy encoder and lightweight decoder format. Keeping our decoder lightweight forces richer embeddings from encoder, suitable for downstream tasks. As mentioned before, we incorporate multiple modalities (spectral imagery and weather) in pretraining our architecture. Figure~\ref{fig:architecture_detailed} shows KG-VSF's architecture. For the forecasting-based pre-training task, it is essential for architecture to capture spatiotemporal nature of satellite data, temporal nature of weather data, and most importantly, create strong embeddings that capture knowledge of causality.

\textbf{Satellite image Encoder/Decoder}:
We use a shared Vision Transformer (ViT) to extract spatial features from spectral imagery across timestamps. ViT have been shown to be effective in the presence of high masking~\cite{he2022masked}, even in geoscience contexts~\cite{jakubik2023foundationmodelsgeneralistgeospatial}. The ViT converts each image into a patch grid of embeddings, incorporating patch positional information. This results in a series of spectral image embeddings on unmasked patches for each timestamp. Since our input is an image series, we propose using a shared ViT across timestamps, leading to a robust encoder, that is capable of embedding images from all timestamps.


\textbf{Weather Encoder}
Due to the coarse spatial resolution of weather data, typically for each image location, we have one weather data point value per timestamp. Thus we use a sequence-to-sequence Unidirectional LSTM \cite{hochreiter1997long} to encode the weather data. The LSTM based approach showed higher accuracy over the transformer based approaches when trying to solely reconstruct masked weather. The LSTM generates weather embeddings for each timestamp, which are then subsampled to match input image timestamps, similar to WSTATT~\cite{wstatt}, called temporal embedding matching. 

\textbf{Timestamp/Delta Encoder}
For temporal information we incorporate day of year (DOY) using a shared linear layer with tanh activation, creating DOY embeddings for each timestamp. Additionally, we create another series that corresponds the number of days in between the images, i.e the delta in timestamps. We generate embeddings for these time delta between images using a separate linear layer, similar to \cite{nguyen2023climax}. The DOY embeddings provide temporal context, while delta embeddings, shown useful in forecasting tasks~\cite{nguyen2023climax}, informs the model about forecast distance.

\textbf{Multimodal Sequence Encoder}
From the previous steps we have spatial, weather, and DOY embedding series. Since all these series are of the same length, we add them all along the temporal dimension to create the multimodal embedding series. We use a transformer with forward-only attention to extract the spatio-temporal information from this multimodal embedding series, analyzing embedding patches from the same spatial location across timestamps. Further we add forward only attention (causal) in transformer, i.e the temporal embeddings created are not bidirectional in nature. This feature makes sure there is no information leakage from future timestamp information to previous timestamps embeddings. The resulting embedding series $Emb_{STW} = [Emb_{:t_1}^{t_1}, \dots, Emb_{:t_C}^{t_C}]$ can be used for downstream tasks, with flexibility in embedding selection. 

\textbf{Forecaster}
\label{sec:Forecaster}
Since our embeddings are constructed using forward-only attention, we use each of the embeddings in $Emb_{STW}$ to forecast an image in its respective future. For each embedding $Emb^{t_i}_{:t_i}$, the forecaster uses the weather embeddings $Emb^{t_j}_{w}$ as well as the the temporal embeddings $Emb^{t_j}_{t}$ and the delta embeddings to generate the embeddings $Emb^{t_j}$ of the forecast timestamp $t_j$. By incorporating weather data leading up to a particular date, we can make more informed estimates about how the land cover might appear, as weather patterns play a significant role in shaping the landscape over time. We combine the four embeddings through addition and feed it to a series of linear and activation layers that forms our forecaster. The objective of these layers is to morph the embeddings from current timestamp to the future timestamp before passing them to the MLP decoder to get the satellite image forecast.

For all but last embeddings in the series $Emb_{STW}$, we forecast the next time-step, i.e t1 of the input series would be used to forecast the image at t2, t2 would be used to forecast t3 and so on (In a similar fashion to the Causal Language Modeling pretraining task used in Large Language Models \cite{achiam2023gpt}). For the last embedding $Emb^{t_C}_{:t_C}$ we forecast the K'th time-step, which is K-C days into the future. This K is a variable and is sampled for each training instance in a batch, thus the name variable-step forecasting (VSF). This method enables the model to predict future images based on current image embeddings and weather information up to the forecast date, acknowledging weather's impact on land cover.

\textbf{Decoder}
Following this stage, we repopulate these embeddings in their respective unmasked positions in the timestamps and zero out the masked patches to pass to the decoder, as done by most methods in transformer based autoencoder methods\cite{jakubik2023foundationmodelsgeneralistgeospatial,he2022masked}. We use a light-weight MLP decoder (similar to the one used in ViT) to ensure that the main focus of the model is to create strong encoder to capture the best information. Similar, to the encoder, we have a shared decoder that performs the operations on each timestamp using the same weights. The decoder maps the embeddings to the spectral image space and reshapes the output to match the required size of the spatiotemporal stack.

\section{Pretraining}
\label{sec:pretrain_phases}
As mentioned before, our proposed pretraining task is forecasting of satellite imagery using multimodal data in a Knowledge Guided fashion, a pretraining task that varies significantly from the traditional single modality reconstruction. Our proposed approach can be trained using the architecture described above in a direct shot, i.e initialise and update all layers at once. However, this may not be the best way to ensure that the $Emb_{STW}$ embeddings capture causal information, as that is the ultimate aim of our proposed approach. To ensure this information capture, we propose a phase wise pretraining process. Specifically we propose 2 phases of pretraining, where the first phase focuses on getting information from each modality and the second phase focuses on capturing the causal relationship between modalities in the embedding.

\textbf{Phase 1: Masked Reconstruction}
Our architecture consists of several components, each serving a distinct purpose. The VIT encoder/decoder maps images to and from embedding space and can be trained separately via masked satellite image reconstruction. Similarly, the Weather encoder embeds weather data and also can be trained using a masked autoencoder approach with a Unidirectional LSTM. These steps yield embeddings for both spectral imagery and weather data. Next, we integrate the multimodal sequence encoder, training it to reconstruct the entire satellite image series by masking embeddings instead of images, strengthening its sequence-learning capability. This establishes a baseline for the $Emb_{STW}$ embedding series (through addition with the weather embeddings). While this series captures spatial and temporal patterns, it lacks capturing causal relationships. To validate this, we use it as a baseline for downstream task fine-tuning. Finally, in Phase 2, we introduce a forecaster to infuse causal relationships.

\textbf{Phase 2: Forecasting}
From section \ref{sec:Forecaster}, the forecaster consists of linear layers, and although simple, its role is crucial: it translates embeddings from the input timestamp space to the forecasted timestamp space. Achieving this transformation within a few layers is challenging, so the model will seek to learn this relationship from other components, which would be provided by the weather data. Using weather data, along with current, timestamp, and delta embeddings, the forecaster generates future embeddings. Given the limited layers, all embeddings adjust to capture relevant cross-modal information for forecasting. For instance, heavy rainfall leads to fuller lakes, while increased sunlight accelerates vegetation growth. Repeating this process over various samples embeds land growth and change dynamics into the model’s representations. Without weather data, the model lacks a relationship to learn, making forecasting significantly harder. This no-weather variant serves as an additional baseline for downstream tasks.

To summarize, our proposed pretraining task is to predict a spectral image in the future (response) using a series of spectral images in the past (context) along with the weather till that future date (query). We call this pretraining task as Knowledge Guided Variable step Forecasting (KG-VSF), where the model is expected to forecast k steps into the future. To ensure that relationship across modalities is captured in the embeddings, we adopt a two phase pretraining process. Our hypothesis is that this extra knowledge infusion would help greatly in downstream tasks that rely on land growth and change dynamics such as crop prediction, land cover land use change, etc.  A schematic of the stagewise pretraining can be found in appendix \ref{sec:pretrain_phases_appendix}. We mask out patches in the spectral imagery and weather data in all the stages to lead to better embeddings, which we describe in a future section(Sec \ref{sec:Masking}). An alternative to our two phase pretraining approach would be a single-phase, multi-objective pretraining that simultaneously optimizes masked reconstruction and forecasting losses which might appear more elegant conceptually. However, such a formulation could be suboptimal for our setting where the goal is to embed causal interactions between driver and response modalities rather than just spatiotemporal correlations. For additional discussion, see Appendix \ref{sec:twophases_comp_appendix}

\section{Dataset Description}
\subsection{Data Sources}
Our spectral imagery data comes from 6 bands of Sentinel-2 imagery~\cite{drusch2012sentinel} and our weather data is from ERA5 land analysis data~\cite{hersbach2020era5}. We chose these two sources due to their temporal resolutions and more importantly their availability globally from 2021. Our choice of six bands: B2 (Blue), B3 (Green), B4 (Red), B8A (Narrow NIR), B11 (SWIR-1), and B12 (SWIR-2) was made for both scientific and practical reasons. Prior studies (e.g.,\cite{jakubik2023foundationmodelsgeneralistgeospatial}, \cite{brown2025alphaearthfoundationsembeddingfield}) have also used these exact bands to demonstrate good performance on a number of downstream tasks ranging from agricultural to climate oriented tasks. Including all 13 bands would more than double the amount of storage space required.  Working with fewer bands improved model training speed, and decreased model architecture size, both of which were critical given our limited computational resources.


For each satellite image instance, we also collect the day of the year it came from, thus forming a series with values from 1 to 365 and a length the same as the number of image instances for that location. Though ERA5 data source consists of various bands that are useful for land cover related tasks we chose 5 bands that are the primary weather drivers, namely (temperature 2m min, temperature 2m max, total precipitation sum, $u$ component of wind 10m, $v$ component of wind 10m). ERA5 data is available at a daily temporal resolution and a spatial resolution of ~11k meters. We download our data through Google Earth Engine\cite{Gorelick2017GEE}.

To summarize, for each location our data sources are:
\begin{itemize}
    \item \textbf{Spectral Imagery Series}: A series of Sentinel2 Imagery of 6 bands. Length of this series depends on coverage of the location.
    \item \textbf{Weather Data Series}: A series of ERA5 Land data of 5 bands , with a series length of 365 (one per day).
    \item \textbf{Day of Year Series}: A series of the day of the year number for each spectral image in the series. The length of this series is same as the spectral imagery series.
\end{itemize}

\subsection{Datasets}
\label{sec:datasets}
We constructed two datasets: a Global Dataset and a US Cropland Dataset to serve different objectives. The Global Dataset includes samples from diverse regions across the world, providing wide spectral variability that benefits tasks requiring generalizable spectral representations. In contrast, the US Cropland Dataset focuses on agriculturally active regions within the United States, enabling the model to better learn patterns specific to crop growth and phenological changes. This division allows us to examine how pretraining on broad versus domain-specific data influences the quality and transferability of learned embeddings across different downstream applications. Below we describe the sampling strategy and distribution for both these datasets: 

\textbf{Global Dataset:} We begin by randomly selecting 10,000 unique land locations across the globe, each covering a 128×128 Sentinel-2  (approximately 1km x 1km area). These 10,000 locations are divided into training, validation, and testing splits in a 60:20:20 ratio, corresponding to 6,000, 2,000, and 2,000 locations, respectively. For each location, all available Sentinel-2 images for the year 2021 are collected—typically ranging from 40 to 70 timestamps depending on regional coverage (e.g., 70+ images in US and Europe and about 40 images in India). For the pretraining phase, we choose an input series length of 6 images, and select a random image after the 6th image as the image to forecast. This requires sequences of 7 timestamps for each sample that can be used for training, validation, or testing. Since a single location can theoretically produce multiple unique 7-timestamp sequences (e.g., for a location with 50 timestamps, one could create 50-choose-7, i.e approximately 99.8 million possible combinations), we subsample a small number of representative temporal sequences per location to maintain diversity while keeping the dataset size tractable. Specifically, for each of the 6,000 training locations, we randomly sample 7-timestamp sequences 10 times, yielding a total of 60,000 possible samples from which 50,000 are randomly chosen for pretraining. Similarly, for each of the 2,000 validation and 2,000 testing locations, we sample 5 7-timestamp sequences per location, resulting in 10,000 samples in each split. This results in an overall 50k/10k/10k split. This approach ensures that while spatial coverage remains balanced across splits, the model is also exposed to rich temporal variability within each region.

\textbf{US Cropland Dataset:} We begin by selecting and downloading 50 Sentinel tiles across the contiguous US for the year of 2021, focusing on tiles that contain a high amount of crop cover. Each Sentinel tile covers a region of 10980x10980 pixels at a 10m resolution (approximately 110km x 110km area). Since our sample analysis size is 128x128, numerous non overlapping 128x128 regions can be selected from the overall 10980x10980 region, specifically approximately 85x85 (=7225) non overlapping 128x128 regions can be created. Also, similar to the Global Dataset, for each tile, all available Sentinel-2 images for the year 2021 are collected and filtered to ensure that their cloud cover/missing values are less than 20\%. Similar to the Global data set, we sample 7-timestamp sequences for each 128x128 region for each of the sentinel tiles. Specifically, we randomly select a 128x128 region from each sentinel tile, and then further identify a unique 7-timestamp sequence for that region to create a sample.  Selection of 128x128 regions are biased towards the regions  that have more crop coverage.  For each tile in our dataset, we create 2000 such samples, resulting in a total of 100000 samples(50 tiles* 2000 per tile). We then choose 30 tiles for training and 10 tiles for validation, and 10 tiles for testing, resulting in a 60k/20/20k data split. This approach ensures that there is spatial coverage across different cropland areas in the US. 

Since each sample from both datasets covers an area of 1.28km x 1.28km(i.e 128 x 128 Sentinel Pixels), The weather data per timestamp will be of shape 1x1, i.e one value per weather band per day. The reason behind having only one pixel value from weather data is due to its coarse resolution (11km) compared to the area coverd by our Sentinel patch (1.28km x 1.28km)

\subsection{Masking}
\label{sec:Masking} 
Since we have a spatiotemporal input and architecture, we adopt a spatiotemporally uniform masking method, i.e, masking that is fair both spatially and temporally. In our unique masking strategy, there are an equal number of masked patches per timestamp as well as an equal number of masked patches per spatial patch location along the temporal axis. This ensures that all temporal patch series that can be created from unmasked patches at each spatial location would be of same length, easing the implementation of temporal components. Also the shared vision transformer will also have same number of outputs per timestamp due to same number of unmasked patches per timestamp. For a visual representation of this masking, please refer to the Appendix Section \ref{sec:masking_appendix}.

\section{Experimental Evaluation}
\subsection{Comparative Pretraining Frameworks}
In our experiments, we evaluate the effectiveness of KG-VSF by comparing with several representative foundation modeling frameworks. We model these frameworks by varying the choices for input and pretraining tasks from existing works, as described below:
\begin{itemize}
    \item \textbf{SM-MR}: Single Modality Masked Reconstruction (SM-MR) with satellite data as input and MR as the pretraining task. This is the most common setup used in existing remote sensing foundation models (e.g.,\cite{jakubik2023foundationmodelsgeneralistgeospatial,cong2022satmae,sun2022ringmo}).
    \item \textbf{MM-MR}: MultiModal Masked Reconstruction (MM-MR) with satellite and weather data as input series and MR as pretraining task. This represents the foundation models that use multiple modalities but still do masked reconstruction (eg. \cite{nedungadi2024mmearth,tseng2023lightweight})
    \item \textbf{SM-VSF}: Single Modality Variable step Forecasting (SM-VSF) with the satellite data as input series and Variable Step Forecasting(VSF) as pretraining task.
    \item \textbf{MM-VSF}: Multi Modality Variable step Forecasting (MM-VSF) with the satellite data and weather data as input series and Variable Step Forecasting(VSF) as pretraining task.
\end{itemize}


It is important to note that the goal of the paper is not to release a new foundation model but show efficacy of our novel pre-training strategy for building foundation models. We are aiming to show how a Knowledge Guided pretraining task embeds causal knowledge information, which leads to superior performance on downstream tasks where capturing this causality matters compared to standard pretraining tasks. As a result, we will not be focusing on comparing ourselves with other existing foundation models directly (which are trained with millions of images and heavy gpu resources \cite{jakubik2023foundationmodelsgeneralistgeospatial,cong2022satmae}) but rather implement their frameworks (architecture design, pretraining task, etc.) on our dataset to ensure fair comparison and convince researchers in the future to use our approach for pretraining of their remote sensing based foundation models. For some of the downstream tasks, we do include task specific models and also comparison with Prithvi EO.2 foundation model \cite{szwarcman2025prithvieo20versatilemultitemporalfoundation}.
Note that in our setup the above frameworks can be implemented by changing the inputs and the loss functions, without significant architecture changes. 
Though MM-MR also uses multi modal data, due to the pretraining task of masked reconstruction, the causal knowledge relationship between modalities is not captured in the embeddings and in turn hurts its performance on downstream tasks. We also expect a similar scenario with the MM-VSF pretraining framework as well.

\subsection{Implementation and Pretraining Details}
\label{sec:implementationdetails}
As mentioned in the Dataset Section \ref{sec:datasets}, For the pretraining phase, we choose an input series length of 6 images, and selected a random image after the 6th image as the final image to forecast. We also use 50\% spatiotemporally uniform masking for both forecasting and reconstruction based pretraining(For reasoning behind this choice of 50\% please refer to Appendix Section \ref{sec:masking_appendix}). We use a patch size of 8 for the vision transformer and a hidden dimension size of 512. This means for each 8x8 patch we would get an embedding of size 512.  Models were trained with a learning rate of 0.0001 on 4 A100 Nvidia GPUs using Adam Optimizer\cite{Kingma2015Adam} and Mean Squared Error loss. A batch size of 128 was used and all frameworks were trained for 800 epochs each. If the framework required a forecasting phase, it was trained during the final 300 epochs of the total 800 epochs.

All models were trained till convergence. MM-MR reached lower mean squared loss compared to SM-MR, and KG-VSF reached a lower mean square error loss compared to SM-VSF and MM-VSF. While the output images produced by SM-MR and MM-MR were quite clean, the images produced by SM-VSF and MM-VSF were blurry. KG-VSF was also able to create clean and consistent images, while showing good forecasting performance. For more details please refer to appendix section \ref{sec:pretraining_results}.

It is important to understand that after pretraining is complete, all methods can give an embedding series for past context(i.e satellite and weather data). So for each downstream task, we add a task-specific decoder head that takes the embedding series as input and produces the required outputs. It is important to note that for all downstream tasks only the final decoder layers that are added for that particular downstream task are fine-tuned, while the layers used to create the embedding series are kept fixed (i.e frozen). This design allows a fair comparison of different pre-training objectives in terms of their ability to produce informative embeddings. Consequently, the performance observed on the downstream tasks reflects how effectively each pretext task captures relevant information from the input modalities during the embedding process. For the global data set, the chances of overlap between data used in any of our downstream tasks and 50,000 locations globally is very slim.  For the US Cropland data set, all the data sets used in downstream tasks only come from the 10 test tiles that were never used for pre-training (and validating) the models for creating embeddings.

\section{Results: Pixel Wise Crop Mapping}
\label{sec:pixelwisecropmapping}
In this section, we explore the performance on the downstream task of pixel wise crop mapping.

\textbf{Dataset and Problem Setting}:
Pixel wise crop type mapping has been a topic of great study in the remote sensing community with numerous applications. In this section, we explore the results of finetuning the models pretrained with the US Cropland Dataset for the task of crop type mapping. For results with the Global Dataset please refer to Appendix section \ref{sec:crop_mapping_appendix}. 
Since we are dealing with the models from the US Cropland Dataset, we chose to finetune the model embeddings with regions in the 10 test tiles. In specific, we use just 400 regions of size 128x128 from the Sentinel Tiles of T11SKA, T16SBF, T15TUH, T13SGT in 2021 for finetuning the model’s embeddings for pixel wise crop type mapping for 10 classes(we chose these tiles due to their rich crop cover, and divided each tile into train, validation and test areas following a strategy similar to WSTATT\cite{wstatt}). Like other works\cite{statt,wstatt,calcrop21}, we get our labels for these regions from the Cropland Data Layer(CDL), an annually released land cover map for the entire continuous US by the USDA\cite{USDA_NASS_CDL_2021}.  For each region we use 10 timestamps of data(approximately biweekly from the months of May to September) and the corresponding ERA5 weather data to create embeddings(Reasoning for choosing only 10 timestamps as opposed to using all year data is present in Appendix section \ref{sec:croptimeframe}). For all the pretraining methods, we keep the layers used for creating the embeddings fixed, and add a decoding head with temporal attention for semantic segmentation. For all methods(KG-VSF, SM-MR, etc.) only the attention mechanism and decoder layers are finetuned using the 400 10-timestamps samples using cross entropy loss for 50 epochs each.  Note that the number of timestamps passed in the downstream task is different from the number passed during pretraining highlighting the temporal flexibility of our architecture. For more architecture and implementation details please refer to Appendix Section \ref{sec:crop_mapping_appendix}.



\begin{table}[t]
\caption{{Comparison on downstream task across 5 seeds of crop mapping across the pretraining tasks, Prithvi and crop type mapping models}}
\begin{tabular}{|ccccc|}
\hline
\multicolumn{5}{|c|}{\textbf{Pixel Wise Crop Type Mapping}}                                                                                                                                                         \\ \hline
\multicolumn{1}{|c|}{\multirow{2}{*}{\textbf{Method}}} & \multicolumn{1}{c|}{\multirow{2}{*}{\textbf{Pretrained?}}} & \multicolumn{2}{c|}{\textbf{Modalities Used}}                          & \multirow{2}{*}{\textbf{Mean F1 Score with Std dev}} \\ \cline{3-4}
\multicolumn{1}{|c|}{}                        & \multicolumn{1}{c|}{}                             & \multicolumn{1}{c|}{\textbf{Satellite}} & \multicolumn{1}{c|}{\textbf{Weather}} &                                        \\ \hline
\multicolumn{1}{|c|}{STATT \cite{statt}}                   & \multicolumn{1}{c|}{No}                           & \multicolumn{1}{c|}{Yes}       & \multicolumn{1}{c|}{No}      & 0.3032 (0.0249)                        \\ \hline
\multicolumn{1}{|c|}{WSTATT \cite{wstatt}}                  & \multicolumn{1}{c|}{No}                           & \multicolumn{1}{c|}{Yes}       & \multicolumn{1}{c|}{Yes}     & 0.3289 (0.0240)                        \\ \hline
\multicolumn{1}{|c|}{Prithvi EO.2 \cite{szwarcman2025prithvieo20versatilemultitemporalfoundation}}            & \multicolumn{1}{c|}{Yes}                          & \multicolumn{1}{c|}{Yes}       & \multicolumn{1}{c|}{No}      & 0.5050 (0.0189)                        \\ \hline
\multicolumn{1}{|c|}{SM-MR}                   & \multicolumn{1}{c|}{Yes}                          & \multicolumn{1}{c|}{Yes}       & \multicolumn{1}{c|}{No}      & 0.6809 (0.0203)                        \\ \hline
\multicolumn{1}{|c|}{MM-MR}                   & \multicolumn{1}{c|}{Yes}                          & \multicolumn{1}{c|}{Yes}       & \multicolumn{1}{c|}{Yes}     & 0.7056 (0.0169)                        \\ \hline
\multicolumn{1}{|c|}{SM-VSF}                  & \multicolumn{1}{c|}{Yes}                          & \multicolumn{1}{c|}{Yes}       & \multicolumn{1}{c|}{No}      & 0.4407 (0.0224)                        \\ \hline
\multicolumn{1}{|c|}{MM-VSF}                  & \multicolumn{1}{c|}{Yes}                          & \multicolumn{1}{c|}{Yes}       & \multicolumn{1}{c|}{Yes}     & 0.5426 (0.0122)                        \\ \hline
\multicolumn{1}{|c|}{KG-VSF}                  & \multicolumn{1}{c|}{Yes}                          & \multicolumn{1}{c|}{Yes}       & \multicolumn{1}{c|}{Yes}     & \textbf{0.7602 (0.0139)}               \\ \hline
\end{tabular}
\label{tab:avgf1_cropland}
\end{table}

\textbf{Performance}
Table \ref{tab:avgf1_cropland} compares the performance of KG-VSF, other pretraining approaches, crop type mapping specific models(STATT\cite{statt}, WSTATT\cite{wstatt}) and Prithvi EO.2\cite{szwarcman2025prithvieo20versatilemultitemporalfoundation} when finetuned for crop mapping task using the 400 samples described above on 400 test samples(100 from each tile). In this table we compare the Average F1 Scores (along with standard deviations) across 10 crop classes for 5 seeds. One can observe that KG-VSF performs the best across all methods and also shows relatively less variability across methods. We can observe that KG-VSF outperforms both MM-MR and MM-VSF even though both methods also have weather as inputs for creating their embeddings, showing these pretraining methods lack capturing causal relationships between modalities leading to their inferior performance compared to KG-VSF. One can also observe that STATT and WSTATT perform poorly, which is due to the low number of samples used for training the model(400 samples). We observed that if you increase the number of training samples used(around 4000), the performance of these models catches up to KG-VSF(WSTATT reaches an F1 score of 0.84, MM-MR reaches 0.89, and KG-VSF reaches 0.92). From the table, we also see that Prithvi performs worse than SM-MR. This could be due to a number of reasons including different resolutions of embeddings, and different pretraining datasets (Prithvi’s pretraining dataset may not be tuned for cropland areas unlike ours which focuses on crop lands). It is also shown that Prithvi just outperforms Unet, which STATT was shown to beat by a good margin\cite{statt}. Overall, we can see that embeddings from KG-VSF can be finetuned for crop type mapping using little samples and obtain good performance relative to other pretraining methods and models. For classwise results of best seed for all models please refer to Table \ref{tab:classwisef1_cropland} in the Appendix.

\section{Results: Soil Moisture Tasks}
In this section, we compare the performance of fine-tuned models on soil moisture related tasks, specifically soil moisture prediction and soil moisture forecasting.

\subsection{Soil Moisture Estimation}
\textbf{Dataset and Problem Setting}:
Estimation of Soil Moisture at current timestamp using radar imagery has been a topic of interest in the remote sensing community\cite{adab2020machine,lee2019estimation,senanayake2021estimating}. Typically, multiple sources of data including satellite data, radar data, weather data, depth data, etc. are used to estimate in-situ measurements from their site where groundtruth readings are available. In our downstream task, we aim to emulate this task and demonstrate how a pretraining task that captures causal relationships in its embeddings is better for the task of soil moisture estimation. In this section, we explore the results of finetuning the models pretrained with the US Cropland Dataset for this task. For results with the Global Dataset please refer to Appendix section \ref{sec:soil_moisture_prediction_appendix}. We focus on 6 Sentinel tiles (T11SKA, T10SEJ, T14SKC, T15TUH, T16SBF, T14RQT) regions (geographic locations shown in Appendix Figure \ref{fig:soil_moisure_regions}) and obtain the corresponding Sentinel2, ERA5 and SMAP (’soil\_moisture\_am’ band) values \cite{Entekhabi2010SMAP} for those regions at their optimal temporal and spatial resolutions. These SMAP values serve as our groundtruth for soil moisture\cite{li2021improved,FangShen2020}, i.e our goal would be to use the Sentinel and ERA5 data to predict SMAP value (’soil\_moisture\_am’). The utility of this task would be in cases where SMAP values are not present or the SMAP satellite fails and we need to estimate the SMAP values for the missing days(this is possible since we would have satellite imagery and weather data for these days). In this section, we explore the results of finetuning the models pretrained with the US Cropland Dataset for this task, for results with the Global Dataset please refer to Appendix section \ref{sec:soil_moisture_prediction_appendix}.

In our problem setting, the input would be a time series of either one or multiple modalities to get embeddings from the respective pretraining objectives. Then using these embeddings we would aim to estimate the soil moisture at that location for each timestamp. We use a timeframe of 6 satellite images here and aim to estimate the soil moisture at each timestamp. Each tile is split into 100 grids and a 60-20-20 train-val-test split is done, ensuring no overlap in regions. Multiple samples are generated from each grid, and after preprocessing, we retain 300 samples(50 from each tile) for training, 150 for validation(25 from each tile) , and 150 for testing(25 from each tile). Models from all 4 other pretraining tasks are finetuned by adding decoding head for prediction. Note that for all methods, layers used to create embeddings are fixed and only the decoder is finetuned using Mean Absolute Error loss. For more architecture and implementation details please refer to Appendix Section \ref{sec:soil_moisture_prediction_appendix}.


\begin{table}[t]
\caption{Comparison of models fine-tuned across 5 seeds from different pretraining tasks on the downstream task of soil moisture prediction. R Square Score for each experiment setting is shown.}
\begin{tabular}{|ccccc|}
\hline
\multicolumn{5}{|c|}{\textbf{Soil Moisture Prediction}}                                                                                                                                                                                          \\ \hline
\multicolumn{1}{|c|}{\multirow{2}{*}{\textbf{Method}}} & \multicolumn{1}{c|}{\multirow{2}{*}{\textbf{Pretrained?}}} & \multicolumn{2}{c|}{\textbf{Modalities Used}}                                   & \multirow{2}{*}{\textbf{R Square Score with Std dev}} \\ \cline{3-4}
\multicolumn{1}{|c|}{}                                 & \multicolumn{1}{c|}{}                                      & \multicolumn{1}{c|}{\textbf{Satellite}} & \multicolumn{1}{c|}{\textbf{Weather}} &                                          \\ \hline
\multicolumn{1}{|c|}{Satellite only}                   & \multicolumn{1}{c|}{No}                                    & \multicolumn{1}{c|}{Yes}                & \multicolumn{1}{c|}{No}               & 0.1249 (0.0532)                          \\ \hline
\multicolumn{1}{|c|}{Weather only}                     & \multicolumn{1}{c|}{No}                                    & \multicolumn{1}{c|}{No}                 & \multicolumn{1}{c|}{Yes}              & 0.4813 (0.0101)                          \\ \hline
\multicolumn{1}{|c|}{Weather + Satellite}              & \multicolumn{1}{c|}{No}                                    & \multicolumn{1}{c|}{Yes}                & \multicolumn{1}{c|}{Yes}              & 0.4906 (0.0110)                          \\ \hline
\multicolumn{1}{|c|}{Prithvi  EO.2 \cite{szwarcman2025prithvieo20versatilemultitemporalfoundation}}                    & \multicolumn{1}{c|}{Yes}                                   & \multicolumn{1}{c|}{Yes}                & \multicolumn{1}{c|}{No}               & 0.3377 (0.0098)                          \\ \hline
\multicolumn{1}{|c|}{SM-MR}                             & \multicolumn{1}{c|}{Yes}                                   & \multicolumn{1}{c|}{Yes}                & \multicolumn{1}{c|}{No}               & 0.1845 (0.0393)                          \\ \hline
\multicolumn{1}{|c|}{MM-MR}                             & \multicolumn{1}{c|}{Yes}                                   & \multicolumn{1}{c|}{Yes}                & \multicolumn{1}{c|}{Yes}              & 0.6039 (0.0148)                          \\ \hline
\multicolumn{1}{|c|}{SM-VSF}                            & \multicolumn{1}{c|}{Yes}                                   & \multicolumn{1}{c|}{Yes}                & \multicolumn{1}{c|}{No}               & 0.2661 (0.0180)                          \\ \hline
\multicolumn{1}{|c|}{MM-VSF}                            & \multicolumn{1}{c|}{Yes}                                   & \multicolumn{1}{c|}{Yes}                & \multicolumn{1}{c|}{Yes}              & 0.7606 (0.0162)                          \\ \hline
\multicolumn{1}{|c|}{KG-VSF}                            & \multicolumn{1}{c|}{Yes}                                   & \multicolumn{1}{c|}{Yes}                & \multicolumn{1}{c|}{Yes}              & \textbf{0.8081(0.0150)}                  \\ \hline
\end{tabular}
\label{tab:soil_pred_cropland}
\end{table}

\textbf{Performance}:
The performance of finetuned models across 5 seeds on this task are presented in Table \ref{tab:soil_pred_cropland}. We include comparison with Prithvi EO.2\cite{szwarcman2025prithvieo20versatilemultitemporalfoundation} and other baseline implementations including a models trained from scratch that uses only weather, only satellite data and one model that uses both sources of data. We can see that KG-VSF outperforms other methods, once again beating both MM-MR and MM-VSF (both methods also have weather as inputs for creating their embeddings), showing that performance gain of KG-VSF is coming from causal capture within the embeddings. One can also observe methods that rely on only satellite imagery(Eg. Prithvi, SM-MR, SM-VSF) perform suboptimally, which can be explained as satellite imagery being less linked to soil moisture compared to weather, however here we can notice that Prithvi performs better than SM-MR. Overall though, we can see the benefit of KG-VSF embeddings in this task as well.

\subsection{Soil Moisture Forecasting}

\textbf{Dataset and Problem Setting:} 
Soil moisture forecasting plays an important role in the remote sensing domain, enabling various applications such as optimizing crop management, planning irrigation schedules, and improving drought monitoring and mitigation strategies \cite{dubois2021short,togneri2022soil}. In this downstream task, we aim to evaluate the ability of model's embeddings to forecast soil moisture estimates. Specifically, we seek to predict soil moisture at a given location at an arbitrary future time based on embeddings from past satellite imagery and weather and weather data up to that future time. The embeddings come from each pretraining strategy. We focus on the same 6 Sentinel tiles (T11SKA, T10SEJ, T14SKC, T15TUH, T16SBF, T14RQT) regions (geographic locations shown in Appendix Figure \ref{fig:soil_moisure_regions}) and obtain the corresponding Sentinel2, ERA5 and SMAP ('soil\_moisture\_am' band) values for those regions at their optimal temporal and spatial resolutions. Our goal is to forecast SMAP ('soil\_moisture\_am' band) values at some point in the future using embedding from past context and weather upto forecast date. In this section, we explore the results of finetuning the models pretrained with the US Cropland Dataset for this task. For results with the Global Dataset please refer to Appendix section \ref{sec:soil_moisture_forecasting_appendix}.

To estimate a soil moisture value in the future, we provide past satellite image time series and weather data to get embeddings from each pretraining objective. Then using those embeddings and weather till a future date we estimate the soil moisture at a future date. We use an input timeframe of 6 satellite images and predict the soil moisture value at a random 7th timestamp in the future. Each tile is split into 100 grids and a 60-20-20 train-val-test split is done, ensuring no overlap in regions. Multiple samples are generated from each grid, and after preprocessing, we retain 300 samples for training, 150 for validation, and 150 for testing. To enhance model generalization, we ensure a diverse distribution of forecast range (i.e. the number of days into the future for which forecasts are made), allowing the model to learn both short-term and long-term forecasting patterns. Since we are forecasting using embeddings from past data and future weather data, we can test this task on all strategies of pretraining.  A decoding head for forecasting is appended and finetuned using Mean Absolute Error loss.  Encoder is kept fixed. For more architecture and implementation details please refer to Appendix Section \ref{sec:soil_moisture_forecasting_appendix}. To reiterate, we use embeddings from each pretraining strategy for capturing past information and use that for forecasting, so the focus of this task is on how the embedding from each pretraining strategy affects in forecasting task.


\begin{table}[t]
\caption{Comparison of models fine-tuned from different pretraining tasks on the downstream task of soil moisture forecasting. Average R Squared Scores of forecasted SMAP values are reported for different forecast day ranges.}
\centering
\begin{tabular}{|c|c|cc|cccc|}
\hline
\multirow{2}{*}{\textbf{Method}} & \multirow{2}{*}{\textbf{Pretrained?}} & \multicolumn{2}{c|}{\textbf{Modalities Used}}              & \multicolumn{4}{c|}{\textbf{Forecast Day Range {\scriptsize(R Squared Scores Reported (higher the better))}}}                                                                                                    \\ \cline{3-8} 
                                 &                                       & \multicolumn{1}{c|}{\textbf{Satellite}} & \textbf{Weather} & \multicolumn{1}{c|}{\textbf{0-25 days}} & \multicolumn{1}{c|}{\textbf{25-50 days}} & \multicolumn{1}{c|}{\textbf{50-100 days}} & \textbf{100+ days} \\ \hline
Weather + Satellite              & No                                    & \multicolumn{1}{c|}{Yes}                & Yes              & \multicolumn{1}{c|}{0.5047}             & \multicolumn{1}{c|}{0.4992}              & \multicolumn{1}{c|}{0.4942}               & 0.4931             \\ \hline
SM-MR                            & Yes                                   & \multicolumn{1}{c|}{Yes}                & No               & \multicolumn{1}{c|}{0.3821}             & \multicolumn{1}{c|}{0.3651}              & \multicolumn{1}{c|}{0.3519}               & 0.3409             \\ \hline
MM-MR                            & Yes                                   & \multicolumn{1}{c|}{Yes}                & Yes              & \multicolumn{1}{c|}{0.5429}             & \multicolumn{1}{c|}{0.5341}              & \multicolumn{1}{c|}{0.4982}               & 0.4465             \\ \hline
SM-VSF                           & Yes                                   & \multicolumn{1}{c|}{Yes}                & No               & \multicolumn{1}{c|}{0.4100}             & \multicolumn{1}{c|}{0.3986}              & \multicolumn{1}{c|}{0.3821}               & 0.3619             \\ \hline
MM-VSF                           & Yes                                   & \multicolumn{1}{c|}{Yes}                & Yes              & \multicolumn{1}{c|}{0.6639}             & \multicolumn{1}{c|}{0.6142}              & \multicolumn{1}{c|}{0.5534}               & 0.5139             \\ \hline
KG-VSF                           & Yes                                   & \multicolumn{1}{c|}{Yes}                & Yes              & \multicolumn{1}{c|}{\textbf{0.7834}}             & \multicolumn{1}{c|}{\textbf{0.7034}}              & \multicolumn{1}{c|}{\textbf{0.6532}}               & \textbf{0.6164}             \\ \hline
\end{tabular}
\label{tab:soil_smap_forecast_cropland}
\end{table}

\textbf{Performance:}
We train a model for each of the pretraining task settings using all tiles and a diverse range of forecast day (i.e varying number of days into the future for forecasting from 5 days all the way upto 170 days). To compare the performance of these models systematically, we compare predictions made across different forecast ranges in the test grids, reported in Table \ref{tab:soil_smap_forecast_cropland}. From the Table, we can see that across all forecast day ranges KG-VSF's finetuned model performs the best. We can that in the initial short day ranges, the performance is quite separated but as we go larger forecast day ranges, we can see that performance saturates. This is caused due to the low number of samples used for finetuning, as the forecast day range increases, the effect of the embedding would diminish and the model would rely on the future weather to predict the soil moisture at that point. Nevertheless, we see that KG-VSF performs the best showing that an embedding that captures causal information is beneficial.

\section{Results: Spectral Imagery Tasks}
In this section, we explore some spectral imagery related downstream tasks, specifically missing image prediction and forecasting of future satellite image. 

\subsection{Missing Image Prediction}
{\textbf{Dataset and Problem Setting}}:
With the vast amount of satellite imagery being captured on a daily basis, corrupted or missing data is an occurring phenomenon. In some cases, the areas of missing/corrupted data is provided as a mask, but in other cases this mask is not provided, making it hard to clean or filter data, thus leaving researchers to use these corrupted images in their work\cite{chen2022joint,zhao2021evaluation}. In this downstream task, we aim to fill these corrupted/missing values in the absence of these masks, i.e the methods have no access/information as to which pixels are corrupted, making it a very challenging task.  It therefore tests whether the pretrained embeddings can infer physically plausible content under uncertainty, a realistic use case for cloud-removal or data-gap filling where clouds obscure parts of the image or no cloud mask is available. This setting is thus more challenging and practically meaningful than the masked-reconstruction tasks where the masked portions (e.g., a cloud mask) are known. In this section, we explore the results of finetuning the models pretrained with the Global Dataset for this task.

In this problem setting, we aim to use a spatiotemporal spectral imagery series in which one or more images have missing data and predict those missing values. For our dataset, we chose to sample a new 1000 locations globally, similar to how our base dataset was created. We then did a 60-20-20 training-validation-test split and used patch series from the 600 patches for finetuning. We ensured that there is no overlap in regions across the 1000 newly sampled patches. We chose a input series length of 6, and varied the amount of missing values in input to make the task more challenging. Due to the prediction related nature of this task, models from all 4 pretraining tasks are finetuned. Decoder is reinitialised and finetuned using Mean Square Error loss. Encoder is kept fixed. For more architecture and implementation details please refer to Appendix Section \ref{sec:missing_image_prediction_appendix}.

\textbf{Performance}:
Table \ref{Tab:mse_miss_img} compares the Mean Squared Errors across the various finetuned models. We can see that as the percentage of missing values increases, the MSE values go up, which can be expected due to the task getting harder. However, we can note that the KG-VSF pretrained models have significantly lower MSE values when compared to MR variants. We also note that KG-VSF performs the best across all missing value percentages, with the error not rising as much as the other variants. This is because the causal nature of KG-VSF helps in filling in the missing values. We also observe that the difference between SM-MR and MM-MR is not very high, thus furthering our hypothesis that MM-MR method of pretraining does not effectively capture the relationship between the different modalities of data.

\begin{figure}[t]
  \centering
  \includegraphics[width=0.9\linewidth]{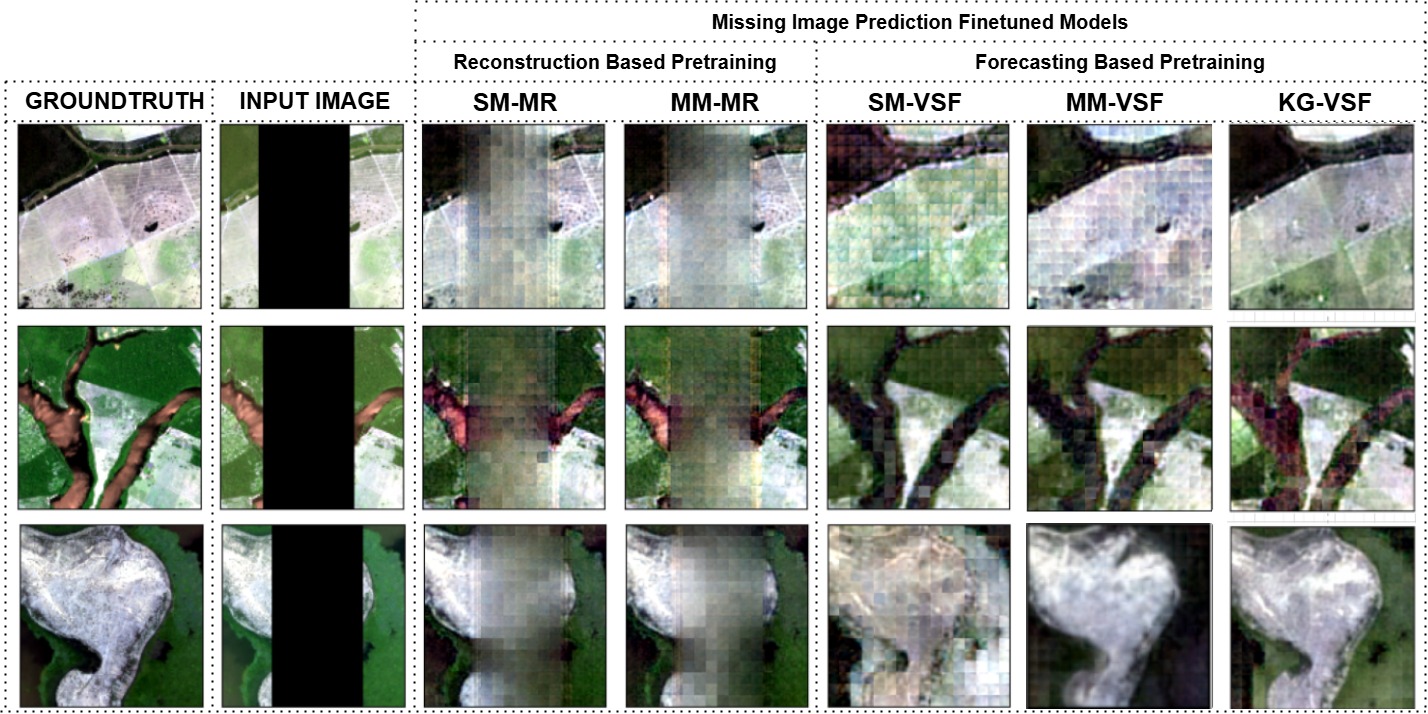}
  \caption{{Comparison of models finetuned from different pretraining tasks on the downstream task of missing image prediction. Predictions for 50\% missing values across finetuned models is shown in the figure above. We can see that the KG-VSF finetuned model estimates the missing values better than other finetuned models.}}
  \label{fig:miss_img_imp_4waycomp}
\end{figure}


\begin{table}[t]
\centering
\caption{\small{Comparison on missing image prediction downstream task across models finetuned from different pretraining tasks. Mean Squared Errors on various levels of missing image percentages shown.}}
\vspace{-0.3cm}
\begin{tabular}{|c|c|cc|ccc|}
\hline
\multirow{2}{*}{\textbf{Method}} & \multirow{2}{*}{\textbf{Pretrained?}} & \multicolumn{2}{c|}{\textbf{Modalities Used}} & \multicolumn{3}{c|}{\textbf{\% Missing Images {\scriptsize(MSE reported)}}} \\ \cline{3-7}
 &  & \multicolumn{1}{c|}{\textbf{Satellite}} & \textbf{Weather} & \multicolumn{1}{c|}{\textbf{50\%}} & \multicolumn{1}{c|}{\textbf{70\%}} & \textbf{90\%} \\ \hline
SM-MR  & Yes & \multicolumn{1}{c|}{Yes} & No  & \multicolumn{1}{c|}{792.68} & \multicolumn{1}{c|}{820.46} & 826.23 \\ \hline
MM-MR  & Yes & \multicolumn{1}{c|}{Yes} & Yes & \multicolumn{1}{c|}{788.94} & \multicolumn{1}{c|}{814.75} & 820.43 \\ \hline
SM-VSF & Yes & \multicolumn{1}{c|}{Yes} & No  & \multicolumn{1}{c|}{362.02} & \multicolumn{1}{c|}{394.32} & 404.32 \\ \hline
MM-VSF & Yes & \multicolumn{1}{c|}{Yes} & Yes & \multicolumn{1}{c|}{347.21} & \multicolumn{1}{c|}{367.82} & 378.13 \\ \hline
KG-VSF & Yes & \multicolumn{1}{c|}{Yes} & Yes & \multicolumn{1}{c|}{\textbf{326.43}} & \multicolumn{1}{c|}{\textbf{337.79}} & \textbf{343.88} \\ \hline
\end{tabular}
\label{Tab:mse_miss_img}
\end{table}

In Figure \ref{fig:miss_img_imp_4waycomp} we compare some output images from our 50 percent missing values experiment across the four methodologies. Each row corresponds to a test sample, and a common trend across rows is that finetuned models pretrained using MR based methodologies do not fill the missing portions very well. In the first row, we can see that SM-VSF has added some false greenness, and MM-VSF does capture greenness, but KG-VSF does not make these mistakes, showing that the embeddings have captured how weather information leads to growth. Row 2 depicts a dried up river bed, but one can observe that SM-VSF and MM-VSF fill the image with a wet river bed, whereas KG-VSF correctly estimates the dried up bed, once again proving embeddings have captured relevant causal satellite-weather information. In the final row, we see a case where SM-VSF and MM-VSF do not green up regions, but KG-VSF did, which once again is a result of causal embeddings.

\subsection{Zero Shot Future Image Forecasting}
\label{sec:Pretrain_results}
\textbf{Dataset and Problem Setting:}  
In this downstream task, we would aim to estimate a satellite image in the future given past satellite and weather data as well as weather up to that future date. Some applications where this task setting is applicable includes estimating how crops will grow in future (use the forecasted growth to calculate crop yield), predicting unavailable image due to revisit interval of satellite (Sometimes satellites do not revisit for upto 15 days, leaving researchers without a way to obtain a good estimate of the imagery in that timeframe), etc. This task requires understanding how weather interacts with land cover or else the models would rely simply on correlation between available past imagery. The pretraining objective of KG-VSF is designed to capture this interaction. This downstream task is actually very similar to our pretraining objective, except here we focus only on the prediction of the last image of the series and do not include any masking of satellite imagery with no finetuning, hence the name zero shot. In this section, we explore the results of the models pretrained with the Global Dataset for this task. Since there is no overlap, we can use the same training, validation and testing regions used in pretraining. Due to the zero shot forecasting related nature of this task, it would be inappropriate to use the SM-MR and MM-MR pretrained models. All pretrained frameworks are kept fixed for the prediction of future image. For more architecture and implementation details please refer to Appendix Section \ref{sec:future_image_forecasting_appendix}.

\textbf{Performance:} Table \ref{Tab:image_forecast} compares the finetuned models performance on the image forecasting task, reporting average mean squared errors for different forecast ranges. Here we aim to predict all 6 of the spectral bands at the future date. We can observe that in the low forecast range (i.e 0 to 25 days into the future), KG-VSF finetuned model performs moderately better than the SM-VSF finetuned counterpart. However, as we increase the forecast day range, we can see the difference in performance between the two models diverge, with significant differences in beyond 100 day forecasts. This can be attributed to the fact that the encoder of KG-VSF has learnt the causal relationship between weather and satellite imagery and so is able to forecast well into the future with minimal finetuning. To highlight this further, some specific examples in the high forecast range are explored.

\begin{figure}[t]
  \centering
  \includegraphics[width=\linewidth]{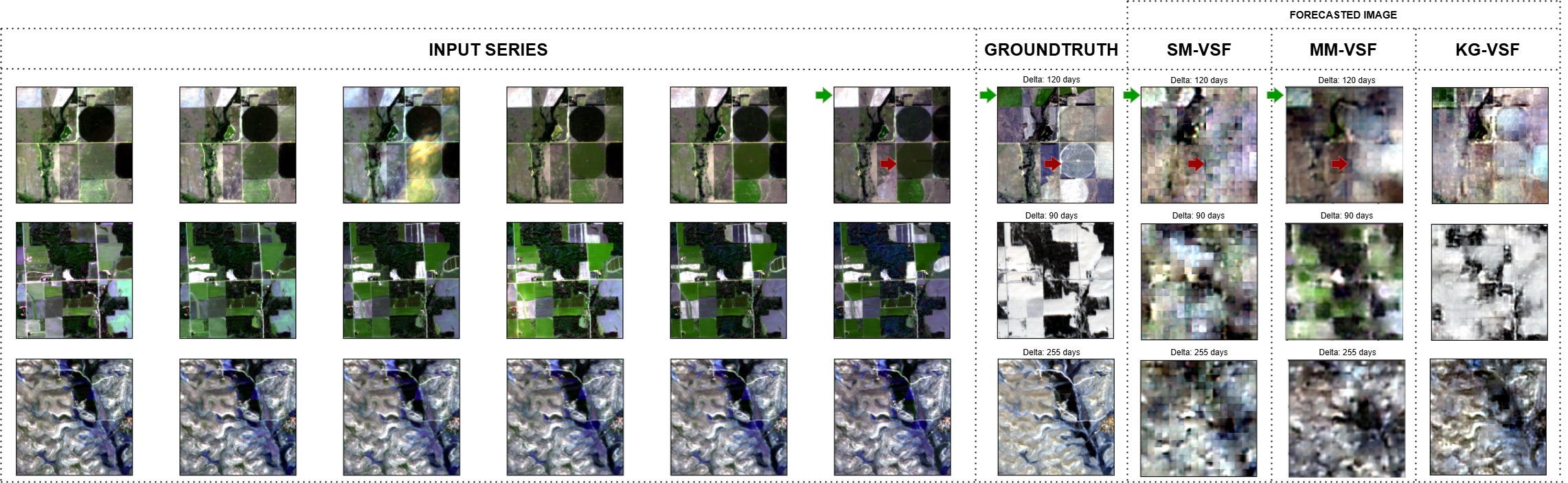}
  \vspace{-0.3cm}
  \caption{{Image Forecast Downstream Task comparison. Row 1 depicts a crop field, Green arrows depict regions of growth, and Red arrows depict regions of harvest, KG-VSF captures both these phenomena better than SM-VSF. Row 2 depicts a case where KG-VSF adds snowfall accurately compared to SM-VSF. Row 3 depicts a case where KG-VSF does not change land cover due to terrain but SM-VSF adds false greenness. Please zoom for better viewing.}}
  \label{fig:forecast_pretrain_comp}
\end{figure}

Figure~\ref{fig:forecast_pretrain_comp} shows a comparison of the forecasted images from these models,i.e SM-VSF, MM-VSF and KG-VSF, on 3 independent examples. Each row correspond to a sample, with the first 6 images corresponding to the satellite component of the input series to the model, the weather component is not shown in the image but is passed along with the satellite component(as shown in Figure \ref{fig:architecture_detailed}) for both MM-VSF and KG-VSF. 

\begin{table}[t]
\centering
\caption{\small{Comparison of models fine-tuned from different pretraining tasks on the downstream image forecasting task. Mean Squared Error (MSE) of the forecasted image is reported for different forecast day ranges.}}
\vspace{-.3cm}
\begin{tabular}{|c|c|cc|cccc|}
\hline
\multirow{2}{*}{\textbf{Method}} & \multirow{2}{*}{\textbf{Pretrained?}} & \multicolumn{2}{c|}{\textbf{Modalities Used}} & \multicolumn{4}{c|}{\textbf{Forecast Day Range {\scriptsize(MSE reported, lower the better)}}} \\ \cline{3-8}
 &  & \multicolumn{1}{c|}{\textbf{Satellite}} & \textbf{Weather} & \multicolumn{1}{c|}{\textbf{0–25 days}} & \multicolumn{1}{c|}{\textbf{25–50 days}} & \multicolumn{1}{c|}{\textbf{50–100 days}} & \textbf{100+ days} \\ \hline
SM-VSF & Yes & \multicolumn{1}{c|}{Yes} & No  & \multicolumn{1}{c|}{340.13}  & \multicolumn{1}{c|}{591.54}  & \multicolumn{1}{c|}{1093.62} & 1112.84 \\ \hline
MM-VSF & Yes & \multicolumn{1}{c|}{Yes} & Yes & \multicolumn{1}{c|}{310.45}  & \multicolumn{1}{c|}{565.21}  & \multicolumn{1}{c|}{1004.21} & 1097.67 \\ \hline
KG-VSF & Yes & \multicolumn{1}{c|}{Yes} & Yes & \multicolumn{1}{c|}{\textbf{237.21}} & \multicolumn{1}{c|}{\textbf{278.83}} & \multicolumn{1}{c|}{\textbf{358.23}} & \textbf{457.27} \\ \hline
\end{tabular}
\label{Tab:image_forecast}
\end{table}

Row 1 depicts an example of a crop field, with the final forecast image being 120 days following the 6th image in the input series. From the groundtruth image, we see that harvest has occurred in the circular fields and growth has happened in the top left corner field. Comparing the forecasted images from SM-VSF, MM-VSF and KG-VSF, we can see that KG-VSF is able to capture both the harvest and the growth of the crops whereas SM-VSF and MM-VSF are not able to capture these changes, showing that causal land cover dynamics have been captured in the embeddings of KG-VSF. Row 2 shows another example of a crop field, with the forecast image being 90 days in the future. We can see from the groundtruth image that 90 days later snow is present in the field, which is captured by KG-VSF but not the others, whose predictions show a faded green field. This illustrates the ability of KG-VSF to capture the relationship between precipitation and temperature on land cover (i.e., precipitation during cold winter days can fall as snow). 
One can also notice that evergreen regions within the forecasted image of KG-VSF have less snowfall when compared to the fields, showing capture of terrain information. This is further reflected in Row 3, where a mountainous region is depicted and the forecast image 255 days in future. We can see that even after 255 days there is not much change in the region, which is correctly captured by our method (KG-VSF), illustrating terrain information capture, whereas SM-VSF seems to add some false greenness. 
We further observe that images from the forecast of SM-VSF and MM-VSF might appear blocky, inspite of long periods of training. This is because the models are unable to forecast images into the future accurately without weather information.


\section{Analysis of Embeddings}
\begin{figure}[t]
  \centering
  \includegraphics[width=0.9\linewidth]{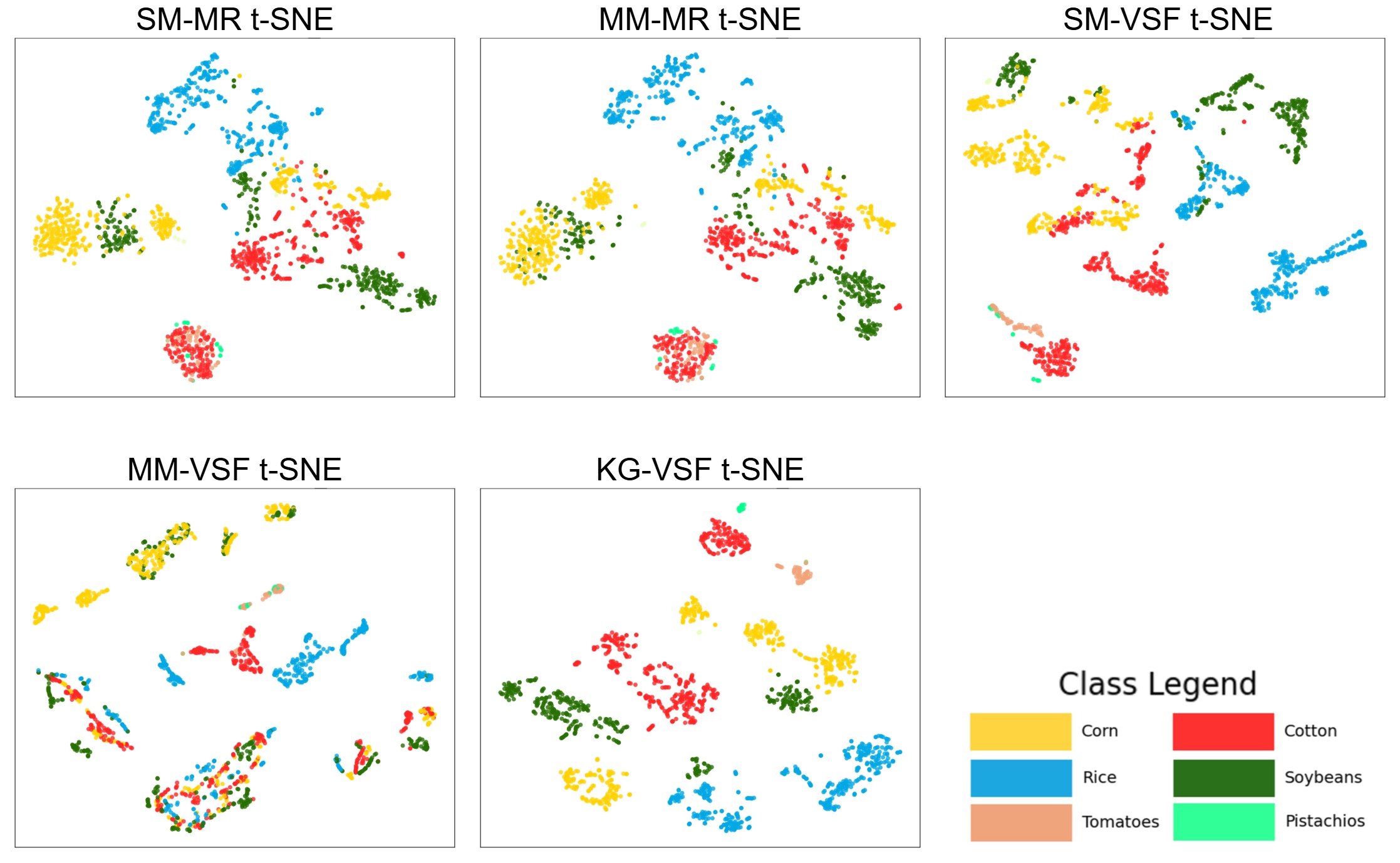}
  \caption{Comparison of embedding quality from the various pretraining objective via t-SNE plots. Embeddings created using satellite and weather data from March to September from specific land cover classes from the various pretraining schemes are projected into 2d space via tsne. One can observe that KG-VSF forms more seperable clusters relative to other pretraining methods.}
  \label{fig:tsne}
\end{figure}

To assess the quality of each pretrained model  we visualize year-end embeddings produced by each pretrained model (SM-MR, MM-MR, SM-VSF, MM-VSF, and KG-VSF) for selected crop types from the US Cropland Dataset for which USDA label quality is known to be generally good. First we resample USDA crop data layer (CDL) labels \cite{USDA_NASS_CDL_2021} for 30-meter Landsat pixels to 10-meter Sentinel pixels\cite{wstatt}. Recall that each 128×128 region is divided into 256 patches each of size 8x8 (section \ref{sec:implementationdetails}).  We then retain only those patches that have at least 80\% purity with respect to a single CDL crop class (i.e., one CDL crop label occurs in at least 80\% of the 64 pixels). The embedding produced for each of these retained 8x8 patches by each pretrained model is projected into two dimensions using t-SNE, and each point is colored according to that patch’s CDL land-cover label. This allows us to directly visualize and compare how well each pretraining approach organizes different land-cover types in its learned latent space in Figure \ref{fig:tsne} (Better models would result in embeddings that create clusters with better separation).

We observe that KG-VSF forms tight, semantically consistent clusters with clear separation between crop types. In contrast, MR-based models produce more diffuse and overlapping clusters, particularly among phenologically similar crops, while VSF variants without knowledge guidance show weaker boundaries.

Overall, these t-SNE visualizations reveal that KG-VSF learns more structured and interpretable representations, organizing patches according to both land-cover semantics and temporal dynamics. This structure emerges purely from the pretrained embeddings, confirming that knowledge-guided pretraining captures meaningful cross-modal relationships even before task-specific fine-tuning.

\section{Do KG-VSF embeddings capture weather-satellite imagery relationship in a physically coherent way?}
To verify that the embeddings produced by KG-VSF are capturing the relationship between weather and satellite images in a physically coherent ways, we performed two experiments. In the first experiment, keeping image history fixed we feed counterfactual weather and observe if the forecasted image changes in the expected direction. In the second experiment, we mute and shuffle weather and observe the performance on downstream tasks of crop type mapping and soil moisture prediction. 

Results of these experiments in the following subsections show that  KG-VSF’s performance gain does not arise from simply having an additional modality at the decoder. Instead, the improvement is due to the model’s ability to leverage physically meaningful weather information during pretraining, resulting in embeddings that encode the true driver-response relationships between weather and surface dynamics.

\subsection{Counterfactual Weather Probing Test}
\begin{figure}[t]
  \centering
  \includegraphics[width=0.9\linewidth]{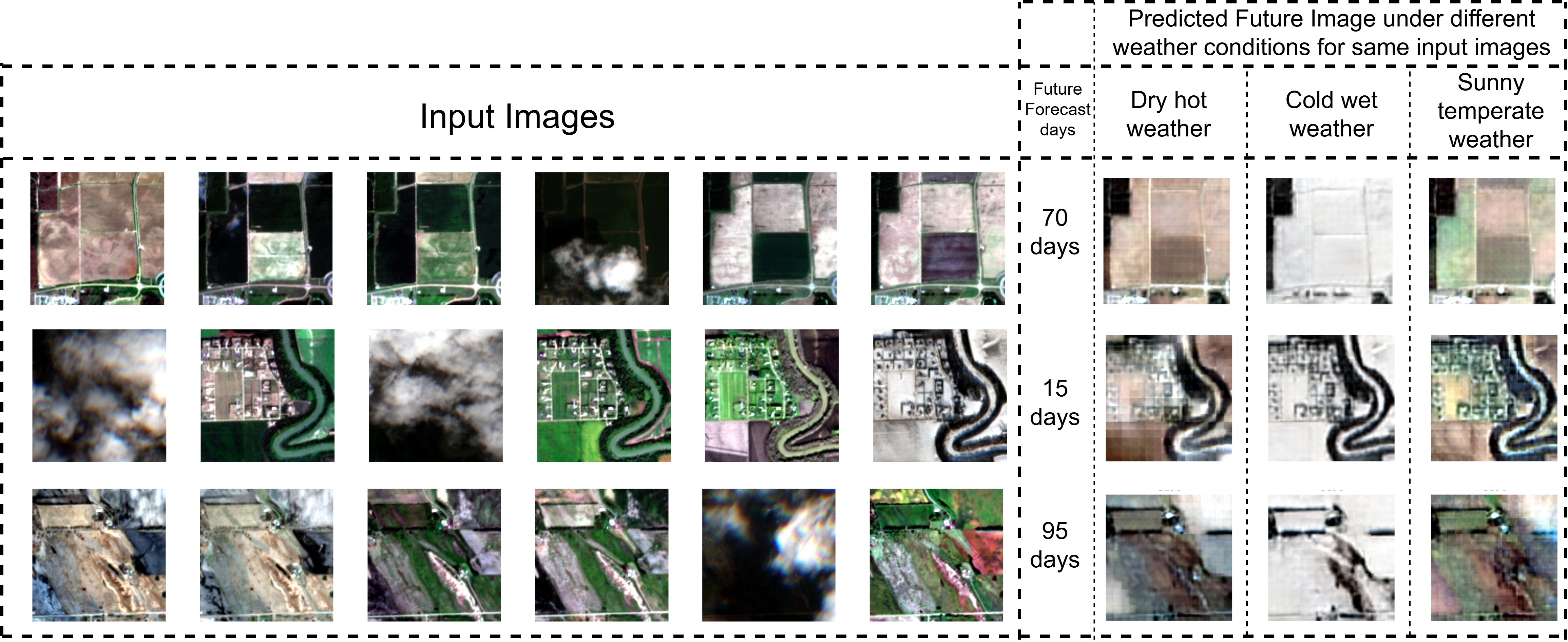}
  \caption{Predicted future images by KG-VSF under different weather conditions (Dry hot, Cold wet and Sunny temperate) for locations in Tennessee (Row 1), MN (row 2 and 3)}
  \label{fig:Causality_check}
\end{figure}

To verify that KG-VSF is capturing causal relations and not just sharpening correlation, in this experiment, we hold the input image history fixed and feed several counter-factual weather sequences and see if the predicted forecast image changes in the expected direction. For example, if weather fed is dry and hot the forecasted image should be dry or if the weather fed is snowy the forecasted image should be snow covered, etc. To conduct this experiment, we chose three samples from the US crop land areas and fed these samples with dry hot weather from Texas (southern latitude), cold wet (snowy) weathers from Northern MN (Northern latitude), and sunny temperate weather from California (middle latitudes) and visualised the outputs produced by model pretrained by KG-VSF in Figure \ref{fig:Causality_check}.

From the figure we can see that under different weather conditions, the forecasted image changes in the physically expected direction. One can notice that under dry hot weather conditions, all forecasted imagery lack greenery. Under the sunny temperate weather condition, forecasted images show greenery. Under the cold wet snowy weather, snow is added to all the images. Note that snow is not added throughout the image but rather added in the places where one would expect it. 
For example, in row 1, snow is added in the open farm land area but not on the parts (left portion of the location) that are covered by trees. Also snow is not added in the green region surrounding the river that may have a slope, making it harder for snow to accumulate there.

\subsection{Mute and Shuffle Weather Test}
We observed from the various downstream tasks that embeddings from KG-VSF show better performance than embeddings from other pretraining schemes. To investigate if  KG-VSF is benefitting from the causal signal itself or simply from having a second modality at the decoder, we conducted an additional ablation experiments on the pixel-wise crop mapping and soil moisture prediction tasks under two settings: (1) weather is muted (ie. all values are made zero) and (2) weather is shuffled as described below:
\begin{itemize}
    \item \textbf{Zeroed-Out Weather:} In this variant, we replace all weather values with zeros, effectively removing the contribution of the weather modality while keeping its structure in the input pipeline. This allows us to verify whether the improvement in the full model arises from the mere presence of a second modality in the decoder. Since the zeroed-out weather input provides no information, any improvement in performance beyond this variant can be attributed to the actual semantic value of the weather data rather than its structural inclusion.
    \item \textbf{Shuffled Weather:} In this setting, we randomly shuffle the weather series across the year for each spatial location. This destroys the temporal and semantic alignment between the weather and the corresponding satellite imagery. As a result, the weather values lose their physical meaning with respect to the visual dynamics in the imagery, preventing the model from leveraging any true causal linkage between the two modalities. Consequently, we expect the embeddings to become suboptimal, leading to degraded performance when fine-tuned for downstream tasks.
\end{itemize}

\begin{table}[t]
\caption{Comparing the performance of shuffling and muting weather experiments on the tasks of crop type mapping and soil moisture estimation. }
\begin{tabular}{|c|c|c|}
\hline
\multicolumn{1}{|l|}{} & Crop Type Mapping & Soil Moisture Prediction \\ \hline
Variant                & Avg F1 Score & R Square      \\ \hline
Zero out weather       & 0.6399       & 0.3143        \\
Shuffle weather        & 0.6933       & 0.3751        \\
Normal                 & \textbf{0.7602}       & \textbf{0.8081}        \\ \hline
\end{tabular}
\label{tab:muteshuffleweather}
\end{table}

Table \ref{tab:muteshuffleweather} compares the performance of these two experiments along with the performance of the base case with correct weather for both pixel wise crop mapping and soil moisture estimation. Across both tasks, we consistently observe that the performance of the shuffled and zeroed-out variants drops substantially compared to the normal case. The shuffled-weather case performs better than the zeroed-out one, likely because it still contains some information about the actual weather (e.g., average temperature, total amount of precipitation). Note that the drop is much steeper for the soil moisture task than for the crop type mapping task. We believe that this is due to the fact that the KG-VSF embeddings are capturing several distinct signals: 1. Patterns in the satellite images, 2. patterns in the weather, and 3. causality between weather conditions and satellite images.  For the crop mapping task, the first component alone likely has a lot of relevant information (as shown in our earlier results on SM-MR) although components 2 and 3 also have additional values.  However the soil moisture depends heavily on the weather (specifically precipitation), and thus muting or shuffling it causes huge degradation in the performance.  


\section{Evaluating Scalability of KG-VSF}
To assess the scalability of our approach, i.e estimate potential downstream gains with larger-scale pretraining, we used the US Cropland dataset spanning 50 Sentinel-2 tiles across the United States as described in Section \ref{sec:datasets}.  As mentioned in Section \ref{sec:datasets}, these tiles were divided into 30 for training, 10 for validation, and 10 for testing, and from each tile, approximately 3,000 spatiotemporal samples were generated using the 7-timestamp Sentinel-2 input configuration.

To systematically explore the impact of scale, we created progressively larger subsets of the training data by sampling an equal number of examples from each of the 30 training tiles: 500 samples per tile (15,000 total), 1,000 per tile (30,000 total), 1,500 per tile (45,000 total), and 2,000 per tile (60,000 total). For each dataset, we trained a model using the KG-VSF pretraining task under identical hyperparameter settings. The resulting pretrained embeddings were then fine-tuned on two representative downstream tasks: pixel-wise crop mapping and soil-moisture prediction.

As shown in Table \ref{tab:scalability}, both downstream metrics improve consistently as the number of pretraining samples increases. This steady upward trend indicates that additional unlabeled Sentinel-2 data would likely yield further performance gains, reinforcing the scalability of the KG-VSF pretraining paradigm. Although computational limitations restricted our experiments to 60,000 samples, the results suggest further improvements are possible with larger datasets that are available publicly from Sentinel-2.

\begin{table}[t]
\caption{Performance of KG-VSF model pretrained with different number of samples on crop type mapping and soil moisture prediction.}
\begin{tabular}{|c|c|c|}
\hline
\multicolumn{1}{|l|}{Scalability Test}                  & Crop Type Mapping & Soil Moisture Prediction \\ \hline
No of samples used in Pretraining KG-VSF & Avg F1 score & R Square      \\ \hline
15000                                   & 0.6496       & 0.6684        \\
30000                                   & 0.7336       & 0.7308        \\
45000                                   & 0.7521       & 0.7889        \\
60000                                   & \textbf{0.7602}       & \textbf{0.8081}        \\ \hline
\end{tabular}
\label{tab:scalability}
\end{table}

\section{Discussion and Conclusion}
In this paper, we introduced Knowledge-Guided Variable-Step Forecasting (KG-VSF), a new pre-training objective for multimodal foundation models that explicitly leverages causal relationships between modalities. Using remote sensing as a case study, we demonstrated that KG-VSF embeddings capture the causal influence of weather (drivers) on land-surface conditions observed via satellite imagery (responses). Compared to traditional pre-training methods such as masked reconstruction and variable-step forecasting, KG-VSF consistently produced embeddings that yielded superior fine-tuned performance across multiple downstream applications, including pixel-wise crop type mapping, soil moisture prediction and forecasting, and spectral imagery-based tasks such as missing-image reconstruction and future-image forecasting.

While KG-VSF has outperformed existing pre-training strategies on downstream tasks used for evaluation in this paper, all of which involve a causal dependency between weather and surface reflectance,  The degree of improvement depends on the strength of this dependency (e.g., this dependency is more direct for the task of image forecasting (section \ref{sec:Pretrain_results}) relative to the task of crop mapping (section \ref{sec:pixelwisecropmapping}).  For tasks that do not have such dependency, KG-VSF may not offer any advantage or may even be worse than existing pre-training strategies. This also suggests that no single pre-training strategy may be universally best in all geo-science applications.

Beyond these examples, the proposed framework opens new possibilities for a wide range of applications where understanding causal interactions among variables is critical. In geoscience, these include wildfire progression forecasting, crop yield prediction, and snowmelt estimation, all of which depend on how environmental drivers influence system responses. More broadly, the same paradigm can be applied to other domains—such as healthcare, where patient outcomes depend on complex interactions among clinical and external factors, or finance, where macroeconomic indicators causally affect market dynamics. By embedding causal reasoning directly into pre-training, KG-VSF provides a principled foundation for developing models that are more physically consistent, interpretable, and transferable across scientific and societal domains.

Looking ahead, we see the main opportunity not merely in scaling up model size, but in expanding the conceptual and methodological space of knowledge-guided pre-training by developing new objectives, architectures, and evaluation frameworks that incorporate domain knowledge in diverse forms. While building full-scale geoscience foundation models will ultimately require substantial computational resources, our results suggest that even modest-scale implementations of knowledge-guided pre-training can yield significant benefits. Given sufficient computational capacity, the KG-VSF paradigm could enable large-scale foundation models that capture causal relationships across diverse domains and data modalities.

\begin{acks}
This research is part of AI-LEAF: “AI Institute for Land, Economy, Agriculture and Forestry,” and is supported by USDA National Institute of Food and Agriculture (NIFA) and the National Science Foundation (NSF) National AI Research Institutes Competitive Award no. 2023-67021-39829. Access to computing facilities was provided by the Minnesota Supercomputing Institute.
\end{acks}

\bibliographystyle{ACM-Reference-Format}
\bibliography{ACM_JOURNAL/acmart-primary/acm_jour}

\appendix

\section{Appendix}

\subsection{Pretraining Stages and Benefit}
\label{sec:pretrain_phases_appendix}
\begin{figure*}[t]
  \centering
  \includegraphics[width=0.8\linewidth]{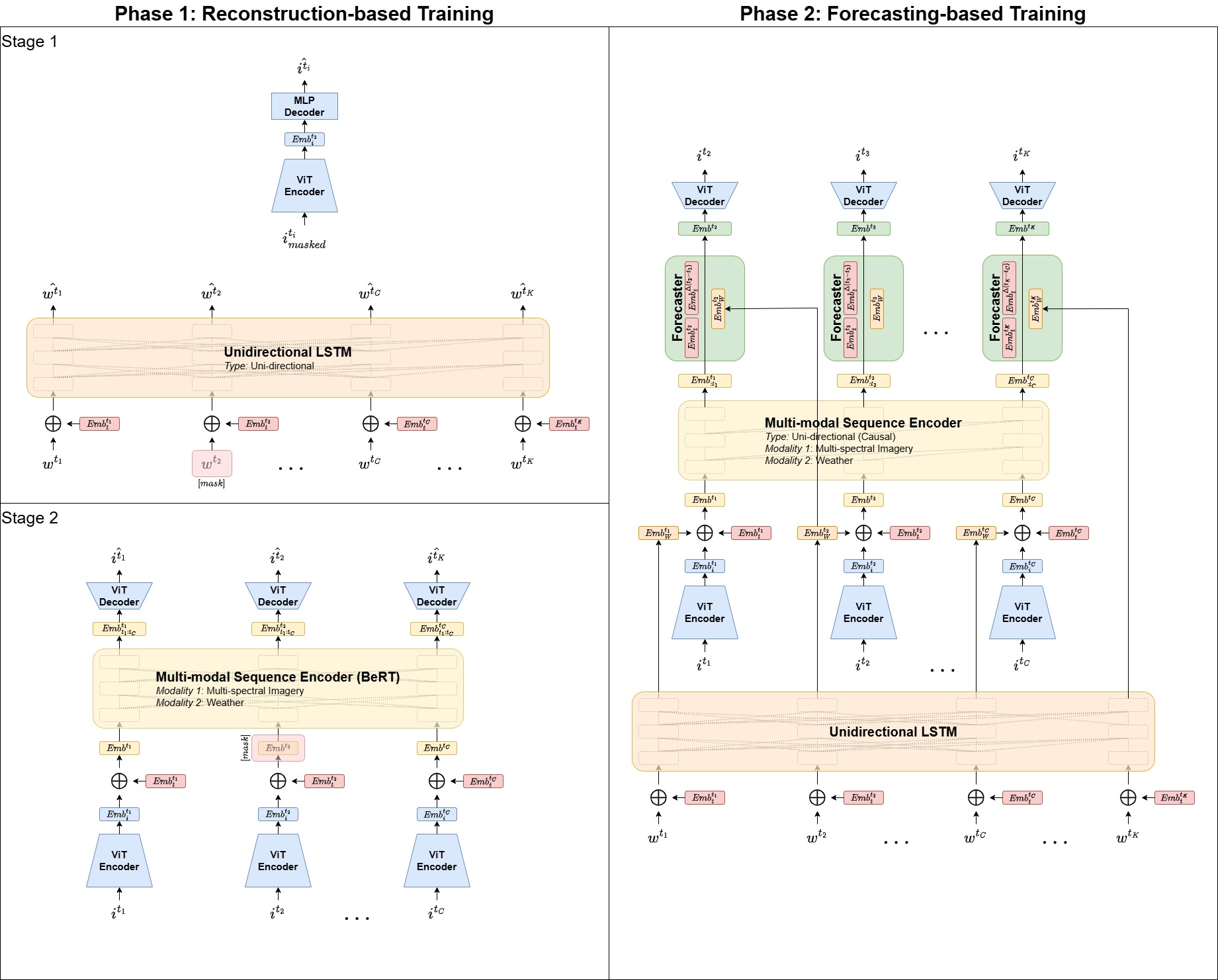}
  \vspace{-0.3cm}
  \caption{\textit{A diagrammatic representation of the proposed stages of pretraining}}
  \label{fig:pretraining_stages}
\end{figure*}

Figure \ref{fig:pretraining_stages} depicts a  diagrammatic representation of the 2 phase stage wise pretraining we propose to ensure that our models's embeddings are able to capture the interaction between modalities. 
\subsubsection{Single-phase vs Two-phase approach for pre-training KG-VSF}
\label{sec:twophases_comp_appendix}
In multimodal contexts like ours, where weather acts as the driver variable and satellite imagery as the response, directly combining masked reconstruction and forecasting objectives in a single loss often can lead to optimization imbalance. The reconstruction objective, being spatially dense and easier to minimize, would tend to dominate the joint loss early in training. As a result, the encoder focuses on learning local texture and spatial smoothness priors from the imagery rather than exploring conditional dependencies across modalities. This yields embeddings that are spatially rich but causally shallow. The forecasting component, which depends on cross-modal reasoning (e.g., inferring how rainfall or temperature anomalies manifest in land cover patterns weeks later), receives weaker gradients during early training, preventing effective modeling of conditional dynamics.

Our two-phase strategy directly addresses this issue. In Phase 1 (masked reconstruction), each modality’s encoder (satellite and weather) and the multimodal sequence encoder are first trained to capture robust representations of spatial and temporal regularities within their respective modalities. This establishes a stable representational foundation that is unbiased toward either input channel. In Phase 2 (forecasting), we introduce a forecaster that must predict future satellite imagery conditioned on both historical satellite context and weather up to the forecast date. Because the reconstruction task is already well learned, the optimization in this phase focuses on aligning weather-driven temporal transitions with satellite responses, thereby embedding causal relationships in the shared latent space.

In summary, while single-phase multi-objective pretraining may be simpler, it tends to emphasize easily optimizable spatial reconstruction signals at the expense of cross-modal causal learning. The proposed two-phase procedure, though slightly more complex, explicitly guides the model from learning modality-specific structure to capturing conditional, knowledge-guided relationships—an essential step for producing generalizable and causally meaningful geospatial embeddings.
To empirically show the benefits of the two phase approach, we compare embeddings from both these models on the pixel wise crop mapping task and soil moisture prediction. From Table \ref{tab:phasewise},  we can see that in both cases, the two phase approach outperformed the single phase multiobjective approach, showing that empirically as well, the two phase approach shows benefit.

\begin{table}[t]
\caption{Empirical Comparison of the two phase vs single phase pretraining approaches}
\begin{tabular}{|l|c|c|}
\hline
                                                               & Crop Mapping & Soil Moisture \\ \hline
                                                               & Avg F1 Score & R Square      \\ \hline
\multicolumn{1}{|c|}{Single Phase Multi objective Pretraining} & 0.7189       & 0.7221        \\
\multicolumn{1}{|c|}{Two Phase wise Pretraining}               & 0.7602       & 0.8081        \\ \hline
\end{tabular}
\label{tab:phasewise}
\end{table}

\subsection{Masking}
\label{sec:masking_appendix}
\begin{figure}[t]
  \centering
  \includegraphics[width=0.9\linewidth]{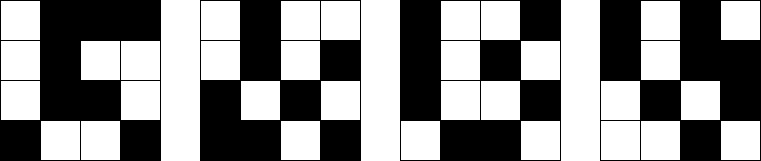}
  \caption{{Example of 50 percent spatiotemporally uniform masking on a 4x4 4 image timeseries}}
  \label{fig:masking}
\end{figure}

Figure \ref{fig:masking} shows an example of such masking for a image series of 4 4x4 grids with 50\% masking. From the Figure, we can see that in each timestamp image there are 8 patches masked and focusing on a particular patch location along the temporal dimension we notice that 2 patches are available. 

\subsubsection{Masking Ratio Comparison}We conduct an experiment to figure out the optimal masking ratio. We train multiple models using 60k samples on different masking ratios and finetuned for the downstream tasks of pixel wise crop mapping and soil moisture estimation. 

\begin{table}[t]
\caption{{Comparison of performance on downstream tasks of crop mapping and soil moisture prediction from KG-VSF models pretrained with different masking ratios}}
\begin{tabular}{|c|c|c|}
\hline
\multicolumn{1}{|c|}{}                   & Crop Mapping & Soil Moisture \\ \hline
Masking Ratio used in Pretraining KG-VSF & Avg F1 Score & R Square      \\ \hline
25\%                                    & 0.7391       & 0.7340        \\
50\%                                    &\textbf{ 0.7602}       & \textbf{0.8081}        \\
75\%                                    & 0.7197       & 0.7689       \\ \hline
\end{tabular}
\label{Tab:maskingratios}
\end{table}

From Table \ref{Tab:maskingratios}, we can see that a moderate amount of masking leads to the best results. This results aligns with the masking results from other papers such as Satmae\cite{cong2022satmae} and ScaleMAE\cite{reed2023scalemaescaleawaremaskedautoencoder}, where they also reported best results when a moderate amount of masking was applied.

\subsection{Pretraining Results}
\label{sec:pretraining_results}
\begin{figure}[t]
  \centering
  \includegraphics[width=0.8\linewidth]{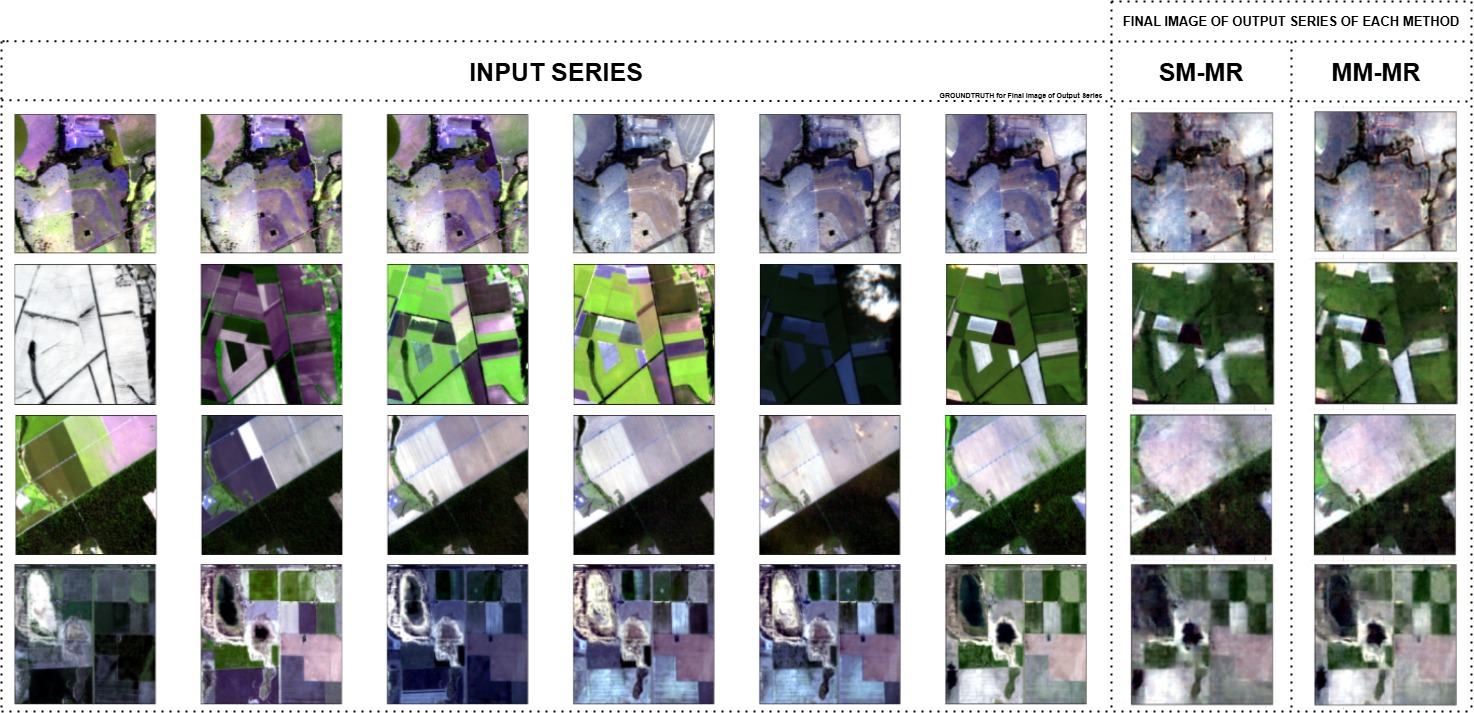}
  \caption{\textit{A comparison of MM-MR and SM-MR on some test samples. As can be seen, MM-MR is slightly better than SM-MR.}}
  \label{fig:pretraining_comp_recon}
\end{figure}

Figure \ref{fig:pretraining_comp_recon} depicts a comparison of SM-MR and MM-MR on some test samples. We can observe that MM-MR does better than SM-MR, but only so slightly. We can notice small differences in each row (lack of greenness, blocky pixels, slight false greenness). Even on comparing Mean Squared error between the two approaches there was not a big difference.

\subsection{Pixel Wise Crop Mapping}
\label{sec:crop_mapping_appendix}

\begin{figure}[t]
  \centering
  \includegraphics[width=0.9\linewidth]{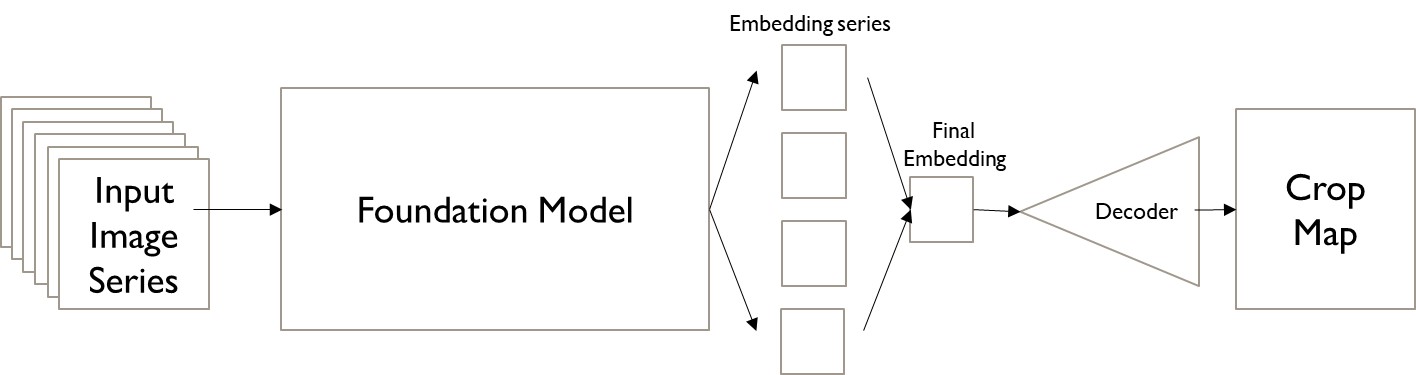}
  \caption{\textit{Layout of architecture for downstream task of crop mapping}}
  \vspace{-0.3cm}
  \label{fig:Crop_mapping_arch}
\end{figure}

\subsubsection{Architecture and Implementation details}:
To perform crop mapping we would need an architecture that gives us a pixel wise output, in particular, we would need a decoder that takes the embedding series given by the pretrained model encoder, i.e $Emb_{STW}$, using it to construct a pixel wise classification map. To map this embedding series to a pixel wise map, we follow an attention based approach, similar to WSTATT\cite{wstatt}. This strategy assigns a weightage to each timestamp and does an aggregated sum to form a multitemporal attention-based embedding. This multitemporal embedding is then acted on by a series of upscaling and convolution layers with activation, and an output Linear layer to form a pixel wise map. Figure \ref{fig:Crop_mapping_arch} depicts the general layout of the architecture used for the crop mapping downstream task. The embedding series mentioned would correspond to the series $Emb_{STW}$, i.e the output series of the encoder.
In all methods(KG-VSF, SM-MR, etc.) the encoder weights are fixed from the respective pretraining task and only the attention mechanism and decoder layers are finetuned. All models finetuning is done with only 2018 data for 40 epochs using Cross Entropy Loss and Adam optimizer with a learning rate of 0.0001.

\subsubsection{US Cropland Dataset Results}:
In Table \ref{tab:avgf1_cropland}, we reported the average F1 score across five seeds for the ten classes of interest. In Table \ref{tab:classwisef1_cropland}, we present the class-wise results for the best-performing seed among the five. Overall, we can see that the KG-VSF finetuned model achieves the best performance across most classes.

\begin{table}[t]
\caption{Comparison on downstream task of crop mapping for the best seed across the pretraining tasks, Prithvi and crop type mapping models}

\begin{tabular}{|c|cccccccc|}
\hline
              & \multicolumn{8}{c|}{2021 Classwise F1 Scores for best model across 5 seeds of testing}   \\ \hline
Model:        & STATT \cite{statt} & WSTATT \cite{wstatt} & Prithvi EO.2 \cite{szwarcman2025prithvieo20versatilemultitemporalfoundation} & SM-MR  & MM-MR           & SM-VSF & MM-VSF & KG-VSF          \\ \hline
Corn          & 0.4768 & 0.4319 & 0.7391  & 0.8461 & 0.8484          & 0.6890 & 0.7626 & \textbf{0.8920} \\
Cotton        & 0.3022 & 0.2989 & 0.7022  & 0.8339 & 0.8290          & 0.6754 & 0.6690 & \textbf{0.8764} \\
Rice          & 0.8048 & 0.7563 & 0.7701  & 0.8827 & 0.9087          & 0.8794 & 0.8833 & \textbf{0.9273} \\
Soybeans      & 0.3541 & 0.2933 & 0.7570  & 0.8932 & 0.9057          & 0.7522 & 0.7238 & \textbf{0.9120} \\
Alfalfa       & 0.0068 & 0.0009 & 0.3405  & 0.6222 & 0.6578          & 0.2903 & 0.3540 & \textbf{0.6673} \\
Tomatoes      & 0.2311 & 0.3380 & 0.4608  & 0.7938 & \textbf{0.8713} & 0.4315 & 0.4697 & 0.8459          \\
Idle Cropland & 0.3161 & 0.4139 & 0.5915  & 0.6087 & 0.5667          & 0.5957 & 0.6719 & \textbf{0.6731} \\
Grapes        & 0.3772 & 0.4987 & 0.1896  & 0.7376 & 0.7331          & 0.0000 & 0.6667 & \textbf{0.7751} \\
Almonds       & 0.4287 & 0.4583 & 0.4146  & 0.4645 & \textbf{0.5686} & 0.1922 & 0.4156 & 0.5569          \\
Pistachios    & 0.0283 & 0.0280 & 0.3251  & 0.3290 & 0.3566          & 0.1663 & 0.0005 & \textbf{0.5860} \\ \hline
Avg           & 0.3326 & 0.3518 & 0.5291  & 0.7012 & 0.7246          & 0.4672 & 0.5617 & \textbf{0.7712} \\ \hline
\end{tabular}
\label{tab:classwisef1_cropland}
\end{table}

\subsubsection{Global Dataset Results}
\begin{figure}[t]
  \centering
  \includegraphics[width=0.8\linewidth]{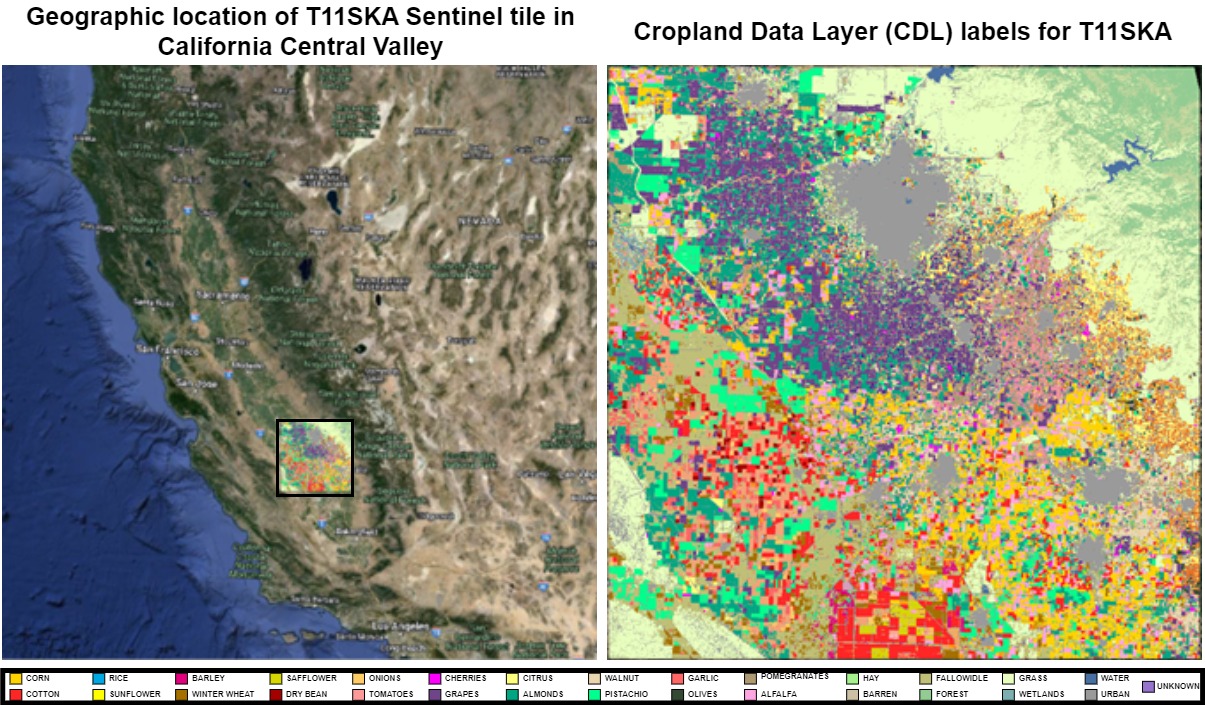}
  \caption{\textit{Geographic location of the T11SKA Sentinel Tile and its corresponding CDL labels. Each color in the CDL image corresponds to a land cover class.}}
  \label{fig:region_of_analysis}
\end{figure}

: 
Our data for finetuning comes from Sentinel2 and ERA5 land data for the region of the T11SKA Sentinel tile in the California Central Valley, a region rich in various crop classes and has been used for crop type mapping in various other works\cite{statt,calcrop21,wstatt} from the years 2018 and 2019. Figure \ref{fig:region_of_analysis} depicts the Cropland Data Layer(CDL) labels and the region of analysis (Sentinel Tile T11SKA) for our crop mapping downstream task. As can be seen our region of analysis lies in the heart of the California Central Valley and contains numerous crop classes. Like other works, we get our labels for this region from the Cropland Data Layer(CDL), an annually released land cover map for the entire continuous US by the USDA. A diagram  of the CDL labels and geographic location of T11SKA tile can be seen in Appendix Figure \ref{fig:region_of_analysis}.

\begin{table}[t]
\centering
\caption{{Comparison on downstream task of crop mapping across the pretraining tasks finetuned using only 2018 data. The models here are pretrained using the Global dataset}}
\begin{tabular}{|c|ccccc|}
\hline
              & \multicolumn{5}{c|}{2019 Test Classwise F1 Scores}                                        \\ \hline
Crop Class    & SM-MR      & MM-MR & SM-VSF       & MM-VSF  & KG-VSF    \\ \hline
Corn          & 0.4135          & 0.4717          & 0.4489  & 0.4836   & \textbf{0.5708} \\
Cotton        & 0.8346          & 0.8403          & 0.9055  & 0.9037   & \textbf{0.9125} \\
Winter Wheat & 0.1156          & 0.0770           & 0.1043  & 0.1129   & \textbf{0.1777} \\
Tomatoes      & \textbf{0.7680} & 0.7637          & 0.7223  & 0.7172   & 0.7341          \\
Grapes        & 0.7398          & 0.7502          & 0.7447  & 0.7355   & \textbf{0.7543} \\
Almonds       & 0.3851          & \textbf{0.4386} & 0.2073  & 0.2509   & 0.2990          \\
Walnut        & 0.0238          & 0.1417          & 0.4494  & 0.4053   & \textbf{0.5384} \\
Pistachio     & 0.5070          & 0.6734          & 0.6200  & 0.6245   & \textbf{0.7003} \\
Alfalfa       & 0.6892          & \textbf{0.7271} & 0.7072  & 0.6973   & 0.7057          \\
Grass         & 0.7760          & 0.8436          & 0.7715  & 0.7831   & \textbf{0.8445} \\
Urban         & 0.6111          & \textbf{0.6408} & 0.6229  & 0.6301   & 0.6191          \\ \hline
Average       & 0.5331          & 0.5789          & 0.5731  & 0.5767   & \textbf{0.6233} \\ \hline
\end{tabular}
\label{Tab:avgf1appendix}
\end{table}

In this problem setting, we use 10 spectral images (biweekly from May to Sept) and weather data (within that timeframe) to get embeddings from each framework and use that embedding series to predict pixel wise crop labels at a 10m resolution. Similar to WSTATT~\cite{wstatt}, we adopt a grid based training method, by splitting the entire region into train, validation and test grids and also follow their preprocessing steps including combination and erosion. Due to the prediction related nature of this task, models from all 4 pretraining tasks are finetuned. A decoding head for semantic segmentation is appended and finetuned using cross entropy loss.  Pretrained Encoder to obtain embeddings is kept fixed. 

Table \ref{Tab:avgf1appendix} compares the performance of KG-VSF and other pretraining approaches when finetuned for crop mapping task. Though finetuning is done using 2018 data, testing is done on 2019 data, assessing the robustness of the approaches. We can see that KG-VSF performs better than other variants. We also observe that MM-MR performs better than SM-MR, which is because inclusion of weather in input is in general better for crop mapping task \cite{wstatt}, however MM-MR and MM-VSF lack capturing the relationship between modalities, thus its inferior performance when compared to KG-VSF. Finally, the ability of KG-VSF’s embeddings to generalize across years after finetuning indicates that the embeddings have captured crucial information that go beyond what is just present in spectral imagery.  Classwise F1 score performance on the test set can be seen in Table \ref{Tab:avgf1appendix}. We can see a good improvement in classes where weather influences crop growth directly, such as Corn and Walnut.


\subsubsection{Comparison of Performance across different timeframes}
\label{sec:croptimeframe}
From the Crop mapping downstream task, we saw we used 10 Sentinel-2 timestamps focused around the growing season of crops (i.e May to to get our results. To ensure that this is the optimal number of  we experimented with multiple temporal variants to assess the model’s sensitivity to the number of observation samples. As shown in Table \ref{tab:croptimeframe}, we can see that as long as the input data is focused on the growing season, the performance remains good. 

\begin{table}[t]
\caption{{Comparison on downstream task of crop mapping across different input timeframes}}
\begin{tabular}{|c|c|}
\hline
\multicolumn{1}{|l|}{}                                                      & Crop Mapping \\ \hline
Input Timeframe                                                             & Avg F1 Score \\ \hline
Growing season(May to Sept) sparse (monthly)                                & 0.7417       \\
Growing season(May to Sept) all timestamps                                  & 0.7237       \\
All year data sparse (monthly)                                              & 0.6734       \\
All year data all timestamps ($\sim$40--60  time steps/year)                & 0.6993       \\
Growing season(May to Sept) frequent (biweekly -10 time steps) (Our choice) & \textbf{0.7602 }      \\ \hline
\end{tabular}
\label{tab:croptimeframe}
\end{table}

\subsection{Soil Moisture Prediction}
\label{sec:soil_moisture_prediction_appendix}
\subsubsection{Architecture and Implementation details}
Since our output is a single value for each timestamp in the series, we use a series of shared linear layers with Relu following each Embedding in the series to get to our output at each timestamp. For each sample, since the images were sampled at random temporally across the entire year the resulting satellite image series has uneven time intervals.  This leads us to also pass the day of year series as input as well. All models were finetuned for 70 epochs. Like the previous task, the encoder weights were frozen and only the shared linear heads were finetuned using Mean Absolute Error Loss and AdamW optimizer with a learning rate of 0.0001. 

\subsubsection{Global Dataset Results}
\begin{figure}[t]
  \centering
  \includegraphics[width=0.7\linewidth]{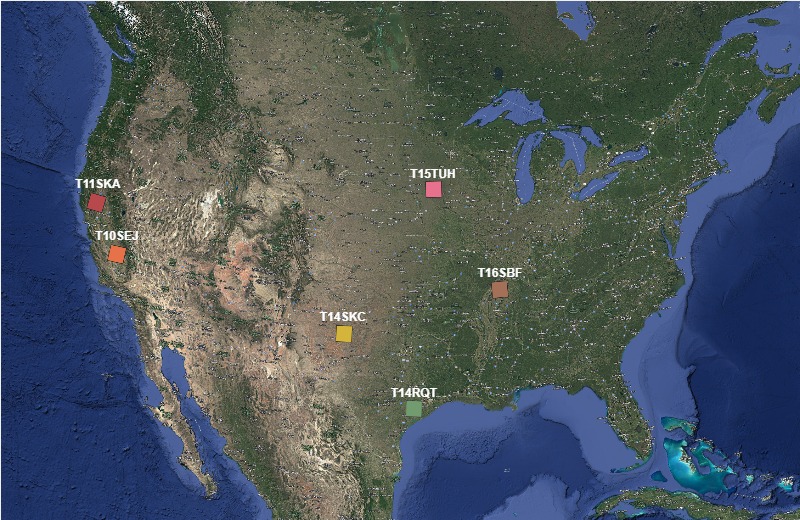}
  \caption{\textit{Geographic location of the 6 Sentinel tiles used in the soil moisture experiments}}
  \label{fig:soil_moisure_regions}
\end{figure}

Figure \ref{fig:soil_moisure_regions} depicts the geographic locations of the 6 tiles used in the soil moisture experiments. From the same 6 Sentinel tiles used in the soil moisture prediction problem, we obtain the Sentinel-2 satellite data and ERA5 weather data to serve as our input modalities and the 'soil\_moisture\_am' band from SMAP to serve as our groundtruth for soil moisture\cite{li2021improved}. For sample generation, like the previous section, each tile is split into 100 grids and a 60-20-20 train-val-test split is done, and after preprocessing, we retain 1800 samples for training (300 from each tile), 600 for validation (100 from each tile), and 600 for testing (100 from each tile). Figure \ref{fig:soil_moisure_regions} depicts the geographic locations of the 6 tiles used in the soil moisture experiments. Due to the prediction related nature of this task, models from all 5 pretraining tasks are finetuned. A decoding head for prediction is appended and finetuned using Mean Absolute Error loss.  Pretrained Encoder to obtain embeddings for past data is kept fixed. 

\begin{table}[t]
\caption{{Comparison of models fine-tuned from different pretraining tasks on the downstream task of soil moisture prediction. Mean Absolute Errors for each experiment setting is shown. In the Test region soil moisture values ranged from 0.0669 to 0.4727 with a mean value of 0.2177 and standard deviation of 0.0972, providing context for interpreting the values below}}
\begin{tabular}{|c|ccccc|}
\hline
                 & \multicolumn{5}{c|}{Soil Moisture Prediction Finetuned Models} \\ \hline
TEST Tile & SM-MR & MM-MR & SM-VSF & MM-VSF & KG-VSF \\ \hline \hline
           \multicolumn{6}{|c|}{In region Testing: Train on 6 tiles, Test on all 6 tiles} \\ 
           \multicolumn{6}{|c|}{\tiny Mean Absolute Error Reported(Lower the better)} \\ \hline
All       & 0.0615 & 0.0458 & 0.0483 & 0.0411 & \textbf{0.0282} \\ \hline \hline
\multicolumn{6}{|c|}{Cross-Region Testing: Train on 5 tiles, Test on the 6th.} \\ 
\multicolumn{6}{|c|}{\tiny Mean Absolute Error Reported(Lower the better)} \\ \hline\hline 
T11SKA    & 0.1113 & 0.0847 & 0.1121 & 0.0913 & \textbf{0.0695} \\ 
T15TUH    & 0.1283 & 0.1365 & 0.1181 & 0.1132 & \textbf{0.0834} \\ 
T14SKC    & 0.1159 & 0.1275 & 0.1312 & 0.1093 & \textbf{0.0958} \\ 
T16SBF    & 0.0821 & 0.0631 & 0.0895 & 0.0619 & \textbf{0.0544} \\ 
T10SEJ    & 0.1003 & 0.0718 & 0.1011 & 0.0742 & \textbf{0.0587} \\ 
T14RQT    & 0.0815 & 0.0579 & 0.0658 & 0.0571 & \textbf{0.0558} \\ \hline
\end{tabular}
\label{Tab:soil_smap_estimation}
\end{table}

The performance of finetuned models on this task are presented in Table \ref{Tab:soil_smap_estimation}. The first experiment conducted was training using samples from all tiles and testing on samples from the test grids of each tile. From this in region sample testing, we can see KG-VSF's finetuned model outperforms other models quite significantly. To explore robustness of learned representations, we also explore cross region testing. In this setting, during finetuning, samples from 1 tile are hidden and during testing, models are tested samples from that hidden tile. For example, models are trained on the 1500 samples from 5 tiles (T10SEJ, T14SKC, T15TUH, T16SBF, T14RQT) and tested on the 100 test samples from the hidden tile (T11SKA). We run this setting hiding each tile one at a time, shown in the lower portion of Table \ref{Tab:soil_smap_estimation}. From these results, KG-VSF's model is able to perform cross region soil moisture estimation better than other models in all tiles, showing that embeddings from KG-VSF are able to capture causality and retain this information even when tested on regions not included in finetuning. This highlights how Knowledge Guided pretraining strategies help in downstream tasks.

\subsection{Soil Mositure Forecasting}
\label{sec:soil_moisture_forecasting_appendix}

\subsubsection{Architecture and Implementation details:}
To capture future weather we use a unidirectional LSTM and add the corresponding hidden state embeddings with the final embedding in the $Emb_{STW}$ series. This multimodal embedding is then used to predict the soil moisture at a future date. Since the output is a single value for each sample time series, we use a series of linear layers with Relu on the final embdeding to lower the dimension to a single value. Since we are finetuning, we freeze the weights of the pretrained encoder and update only the weights of the appended decoder layers and the weather LSTM for 50 epochs using Mean Absolute Error Loss and AdamW Optimizer with a learning rate of 0.0001.

\begin{table}[t]
\centering
\caption{{Comparison of models fine-tuned from different pretraining tasks on soil moisture forecasting downstream task. Table reports average R Squared Score of forecasted SMAP values for different forecast day ranges. In the test region soil moisture ranged from 0.062 to 0.4998 with a mean of 0.2134 and standard deviation of 0.0997}}
\begin{tabular}{|c|c|c|c|c|c|}
\hline
\multicolumn{6}{|c|}{Soil Moisture Forecasting Downstream Task} \\ 
\multicolumn{6}{|c|}{\scriptsize(R Squared Scores Reported (higher the better))} \\ \hline
Future Forecast Day Range & SM-MR & MM-MR & SM-VSF  &  MM-VSF  & \textbf{KG-VSF }      \\ \hline
0 - 25 days         & 0.7231  & 0.8436  & 0.7349 & 0.8607  & \textbf{0.9501} \\ \hline
25 - 50 days        & 0.7098  & 0.7724  & 0.7403 & 0.8019 & \textbf{0.9332} \\ \hline
50 - 100 days       & 0.5818  & 0.7619  & 0.6114 & 0.7735 & \textbf{0.9216} \\ \hline
More than 100 days  & 0.5136  & 0.7593  & 0.5864 & 0.7653 & \textbf{0.9012} \\ \hline
\end{tabular}
\label{Tab:soil_smap_forecast}
\end{table}

\subsubsection{Global Dataset Results}
Figure \ref{fig:soil_moisure_regions} depicts the geographic locations of the 6 tiles used in the soil moisture experiments. Here too, To estimate a soil moisture value in the future, we provide past satellite image time series and weather data to get embeddings from each pretraining objective. Then using those embeddings and weather till a future date we estimate the soil moisture at a future date. We use an input timeframe of 6 satellite images and predict the soil moisture value at a random 7th timestamp in the future. Each tile is split into 100 grids and a 60-20-20 train-val-test split is done, ensuring no overlap in regions. Multiple samples are generated from each grid, and after preprocessing, we retain 1800 samples for training, 400 for validation, and 400 for testing.

From Table \ref{Tab:soil_smap_forecast}, in short forecast ranges (i.e less than 25 days) most finetuned models are able to forecast soil moisture quite well with a moderate difference between them, however KG-VSF is top performing. However, when we move to larger forecast ranges, we can see a big jump in the performance of KG-VSF finetuned model compared to other finetuned models. Even though, as the forecast range increases, errors of all models go up, the errors by the other model go up significantly higher compared to KG-VSF's model. This can be attributed to the fact that the embeddings from the KG-VSF model are able to capture the causal relationships between weather satellite imagery better than the other counterparts, leading to a performance boost.

\subsection{Missing Image Prediction}
\label{sec:missing_image_prediction_appendix}
\subsubsection{Architecture and Implementation details}
Since our output is a series of images, we can use the same architecture as the pretraining model. However, we freeze the encoder and reinitialise the decoder weights to random values and update only these layers. To simulate missing data, we would zero out large blocks from the image and pass an input series with the blacked out images to the model. For our study, we chose to blacken out 2 images per series, both with a particular percentage of missing values. Consequently, we would pass a series of input images with one or more timestamps having missing data, and we would ask our architecture to reconstruct the entire series, but take only the mean squared loss on the timestamps with missing values, using those loss values for backpropagation. Models were finetuned for 30 epochs with Adam Optimiser and a learning rate of 0.0001.

\subsection{Future Image Forecasting}
\label{sec:future_image_forecasting_appendix}
\subsubsection{Architecture and Implementation details:}
Since we are forecasting imagery in the future, we can keep the overall architecture the same as during the pretraining (Figure \ref{fig:architecture_detailed} for the KG-VSF variant) and load the pretrained weights and freeze them. We used a input series of length 6 and did not use masking during this downstream task.


\end{document}